%% file: main.tex
\documentclass[10pt,twocolumn,letterpaper]{article}

\usepackage{cvpr}

\input{preamble}

\def\paperID{2523} 
\def\confName{CVPR}
\def\confYear{2026}

\title{\em{SCP}: Spatial Causal Prediction in Video}

\author{
Yanguang Zhao$^{1}$ \quad
Jie Yang$^{1}$ \quad
Shengqiong Wu$^{1}$\thanks{Corresponding author: Shengqiong Wu.} \quad
Shutong Hu$^{1}$ \quad
Hongbo Qiu$^{2}$ \quad
Yu Wang$^{3}$ \\
Guijia Zhang$^{2}$ \quad
Tan Kai Ze$^{1}$ \quad
Hao Fei$^{1}$ \quad
Chia-Wen Lin$^{4}$ \quad
Mong-Li Lee$^{1}$ \quad
Wynne Hsu$^{1}$ \\
\textnormal{$^1$National University of Singapore} \qquad \textnormal{$^2$Shenzhen University} \qquad \\
\textnormal{$^3$Sichuan University} \qquad 
\textnormal{$^4$National Tsing Hua University} \\
{\tt yanguangzhao@u.nus.edu} \quad {\tt swu@u.nus.edu}
}

\begin{document}
\maketitle

\begin{abstract}
Spatial reasoning, the ability to understand spatial relations, causality, and dynamic evolution, is central to human intelligence and essential for real-world applications such as autonomous driving and robotics. 
Existing studies, however, primarily assess models on visible spatio-temporal understanding, overlooking their ability to infer unseen past or future spatial states. In this work, we introduce \textbf{Spatial Causal Prediction (SCP)}, a new task paradigm that challenges models to reason beyond observation and predict spatial causal outcomes. 
We further construct \textbf{SCP-Bench}, a benchmark comprising {2,500} QA pairs across {1,181} videos spanning {diverse viewpoints, scenes, and causal directions}, to support systematic evaluation.
Through comprehensive experiments on {23} state-of-the-art models, we reveal substantial gaps between human and model performance, limited temporal extrapolation, and weak causal grounding. 
We further analyze key factors influencing performance and propose perception-enhancement and reasoning-guided strategies toward advancing spatial causal intelligence.
The project page is \url{https://guangstrip.github.io/SCP-Bench/}.
\vspace{-5mm}
\end{abstract}

\makeatletter
\let\origaddcontentsline\addcontentsline
\renewcommand{\addcontentsline}[3]{}
\makeatother

\vspace{-3mm}
\section{Introduction}
\vspace{-1mm}
\label{sec:intro}

\begin{figure}[!t]
\centering
\includegraphics[width=0.99\linewidth]{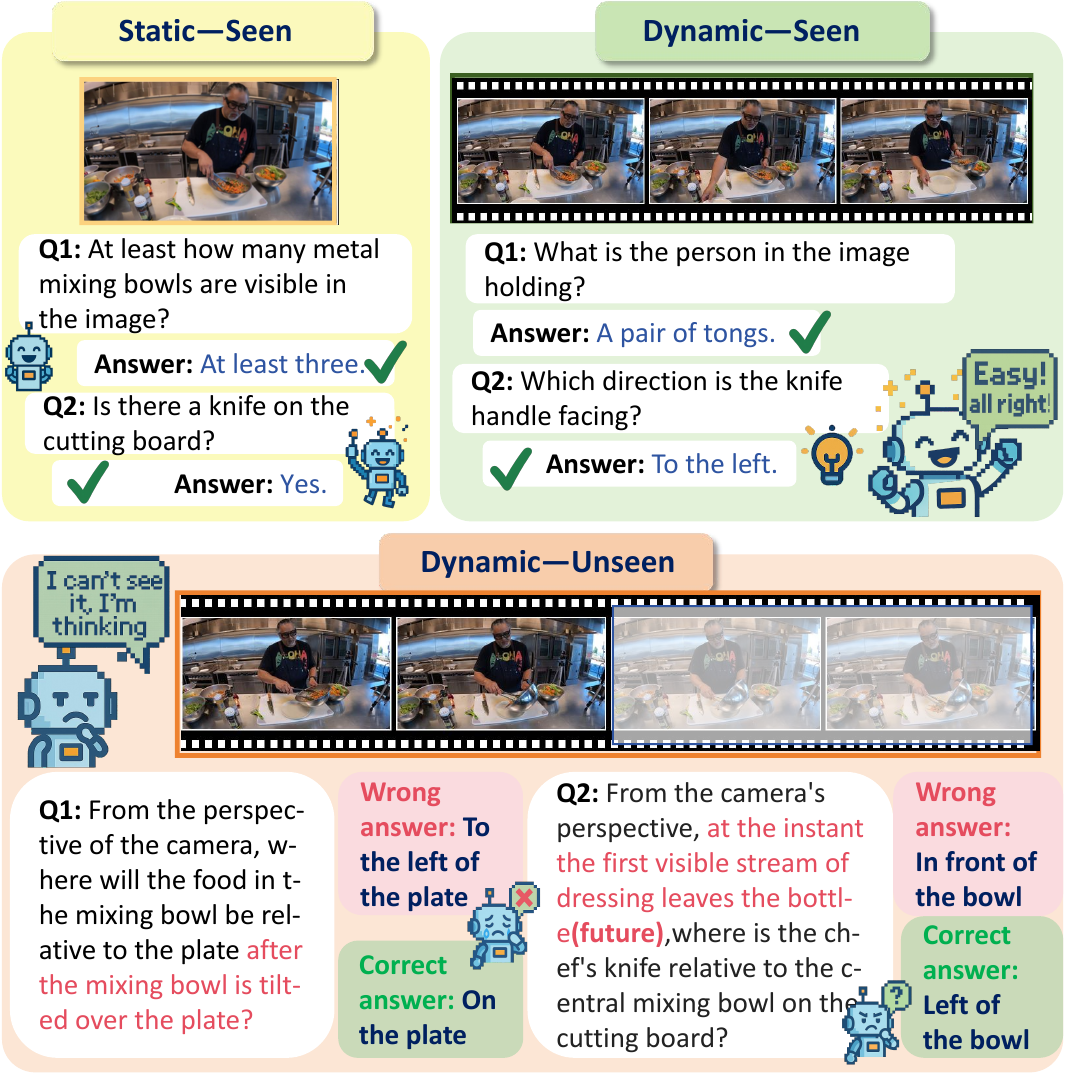}
\vspace{-2mm}
\caption{
Existing benchmarks primarily assess \textbf{known static} or \textbf{known dynamic} reasoning based on fully observable scenes.
A more challenging \textbf{dynamic–unseen} setting is to evaluate models' ability to predict spatial outcomes from partial observations.
}
\label{fig:intro}
\vspace{-7mm}
\end{figure}

Spatial reasoning is a fundamental component of human intelligence that underpins the physical world understanding, enabling perception of spatial relations among objects, the grasp of causality and continuity, and the prediction of future spatial changes~\cite{byrne1989spatial}.
Endowing systems with spatial understanding allows them not only to perceive the visible scene but also to understand the underlying structural regularities and dynamical physical laws, an ability crucial for a wide range of real-world applications such as autonomous driving~\cite{gao2024survey, chen2019autonomous} and robotics~\cite{cai2025spatialbot, song2025robospatial, zhou2025roborefer}.
Prior research has made significant progress toward this goal, with early work primarily focusing on static spatial reasoning that assesses a model's understanding of object layouts and spatial relations from single or multiple viewpoints~\cite{yang2025thinking, deng2025internspatial, yeh2025seeing, wang2025spatial457, jia2025omnispatial, zhang2025mllms, yin2025spatial}.
More recent work~\cite{li2025sti, zhang2025dsi, zhou2025vlm4d} has extended this to spatio-temporal reasoning by introducing videos or dynamic sequences.
However, existing efforts~\cite{yang2025thinking, yu2025far,yang2025mmsi,ma20253dsrbench,wu2025spatialmllm} remain largely limited to perceptual understanding of observed scenes, assessing whether models can reason about visible spatial attributes (e.g., object size, position, or relations) within given inputs, as shown in Fig.~\ref{fig:intro}.
In contrast, real-world spatial reasoning requires the ability to predict future spatial evolution and infer latent causal relations from dynamic observations, beyond merely interpreting what is already visible.

To advance research on causal and temporally aware spatial reasoning beyond the observed scene, we introduce a new task paradigm, \textbf{\textit{Spatial Causal Prediction} (SCP)}, which addresses the limitations of existing studies confined to visible spatio-temporal understanding.
Building on this formulation, we further introduce \textbf{\textit{SCP-Bench}}, a benchmark designed to systematically evaluate a model's capacity to perceive, reason, and predict spatial causal dynamics under partial observability.
We curate various video sources from public datasets and platforms~\cite{abu2016youtube,caba2015activitynet,grauman2024ego,perrett2025hd}, covering diverse \textbf{viewpoints} (\textit{egocentric}, \textit{exocentric}, and \textit{hybrid perspectives}), and \textbf{scenarios} (e.g., \textit{sports}, \textit{driving}, and \textit{factory/machine operations}).
Such a design allows SCP-Bench to assess not only visual perception but also a model's grasp of physical commonsense, causal continuity, and dynamical regularities~\cite{yi2019clevrer, riochet2018intphys, bakhtin2019phyre,azzolini2025cosmos}.
To construct high-quality annotations, we develop a semi-automatic labeling pipeline combining model-assisted generation~\cite{openai2025gpt5} with human verification. Following a QA-based task format, we define {8} \textbf{spatial reasoning categories} (e.g., \textit{relation}, \textit{planning}, \textit{relative size/speed/distance}, \textit{spatial state}), spanning \textbf{two causal directions}, i.e., \textit{backward} (cause inference) and \textit{forward} (result prediction).
After rigorous manual validation, the final benchmark comprises \textbf{2,500} question–answer (QA) pairs across \textbf{1,181} high-quality video clips.
With SCP-Bench, we further investigate three key research questions through extensive experiments and in-depth analyses.

\vspace{1mm}
\begin{infobox}
\textbf{\em RQ1 (\S\ref{How Well Do Current Models Perform}): How do current MLLMs perform on SCP?}
\end{infobox}
\noindent We evaluate 23 representative state-of-the-art MLLMs~\cite{openai2025gpt5,anthropic2025claudeSonnet4_5,comanici2025gemini,an2025llava,ouyang2025spacer,wang2025internvl3,yu2025minicpm, liu2025nvila, wu2024deepseek, wu2025spatialmllm, yang2025qwen3, xu2025qwen3}  spanning both open-source and closed-source families under diverse task types, scenes, and settings.
Our study yields several striking observations.
\textbf{First}, models~\cite{ouyang2025spacer,wu2025spatialmllm} specifically trained for spatial perception often underperform compared to general-purpose models without such specialization, exposing a mismatch between supervised spatial learning and causal generalization.
\textbf{Second}, the performance gap between long-term and short-term spatial causal prediction is surprisingly small, indicating that current architectures offer limited temporal extrapolation benefits.
\textbf{Third}, compared with human reasoning, existing models fall short by roughly 22.37\% in unseen spatio-temporal causal prediction, underscoring the significant gap in causal understanding.
\textbf{Fourth}, large open-source models with increased parameter scales achieve performance comparable to several closed-source systems on SCP-Bench, reflecting the rapid advancement of publicly available spatial reasoning models.

\vspace{1mm}
\begin{infobox}
\textbf{\em RQ2 (\S\ref{sec:affects}): Then, what affects their performance?}
\end{infobox}
\noindent To answer this, we design a series of probing tasks to analyze characteristic error patterns and reasoning behaviors.
Our results show that current models struggle not only with the spatial perception of visible content but also with predicting unobserved spatial states.
On the \textit{perception} side, understanding dynamic video content remains notably more difficult than handling static visual inputs.
Variations in \textit{causal structure}, \textit{motion dynamics}, and the \textit{continuous evolution of spatial states} often disrupt spatial consistency and result in fragile reasoning~\cite{zang2023discovering, foss2025causalvqa}.
On the \textit{prediction} side, many models lack a solid grounding in physical commonsense, which substantially hinders accurate forecasting of unseen spatial configurations~\cite{bisk2020piqa, chow2025physbench}.

\begin{infobox}
\textbf{\em RQ3 (\S\ref{sec:Improvement}): How to improve their SCP capabilities?}
\end{infobox}
\noindent We further investigate several approaches and paradigms for SCP models.
Our experiments reveal that significantly scaling model size consistently yields substantial performance gains.
Moreover, we demonstrate that enriching spatial perception with intermediate representations, such as dense video captions or spatial-interaction graphs, provides the predictor with more enriched scene information, leading to slightly improved SCP performance.
In addition, incorporating prior knowledge of physical commonsense derived from large language models~\cite{openai2025gpt5} or learned world models~\cite{wan2025wan} significantly boosts prediction for unseen spatial states.
Interestingly, both explicit chain-of-thought (CoT)~\cite{wei2022cot,fei2024video} prompting and implicit self-thinking~\cite{guo2025deepseekr1,huang2025visionr1,wang2025multimodal} strategies offer limited improvements.

Overall, this work contributes by
\circnum{1} for the first time formulating the Spatial Causal Prediction (SCP) task that advances spatial intelligence from observed spatio-temporal understanding to inference over unseen past and future.
\circnum{2} We develop SCP-Bench, a high-quality benchmark designed to rigorously evaluate spatial reasoning across two causal directions, diverse scenes and viewpoints, and 8 spatial question categories.
\circnum{3} We perform extensive evaluation and in-depth analysis of existing models on SCP-Bench, uncovering when and why they fail and how to improve them, thereby paving the way for follow-up research.

\begin{figure*}[!t]
\centering
\includegraphics[width=0.99\linewidth]{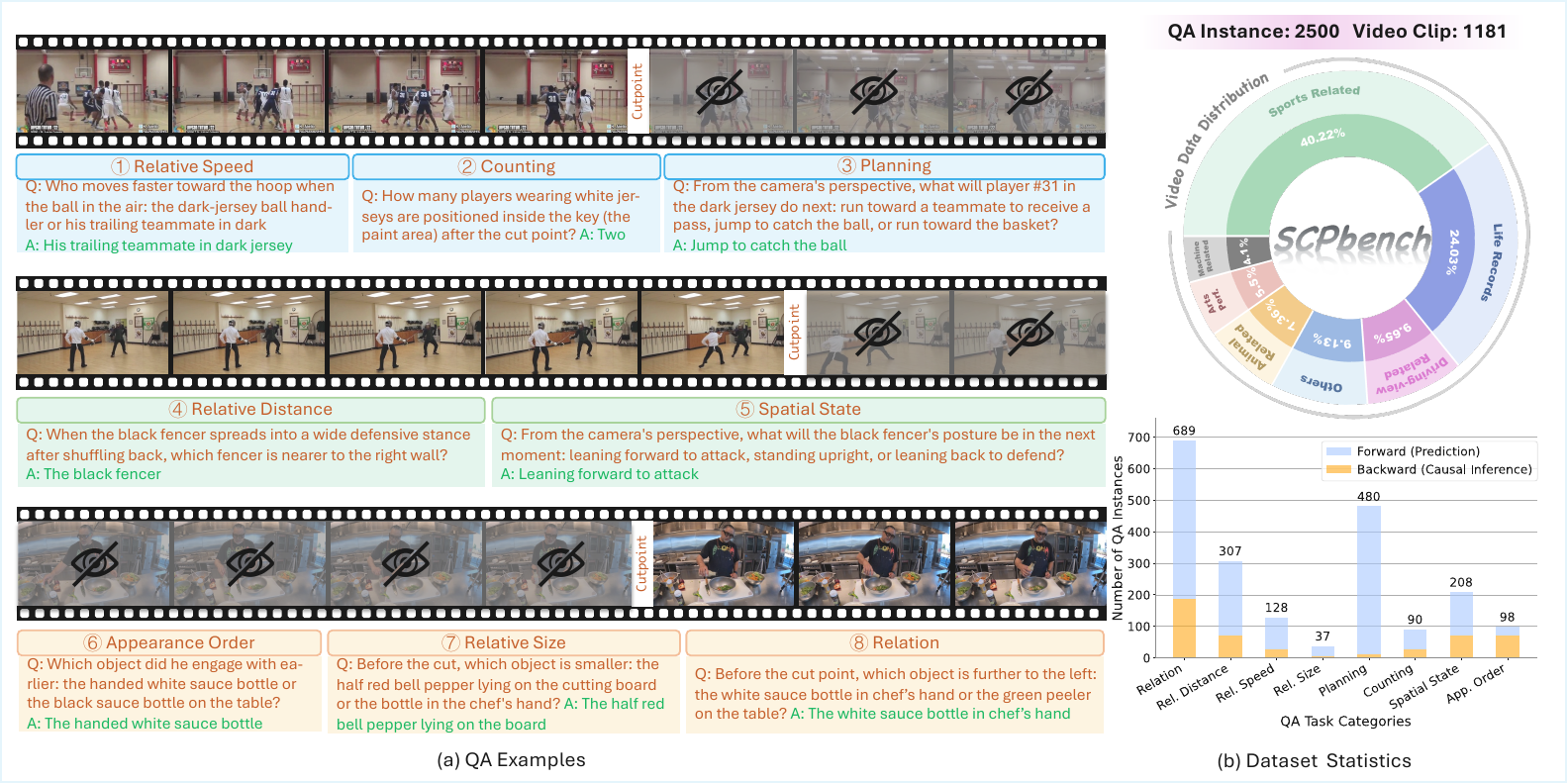}
\vspace{-2mm}
\caption{Overview of SCP-Bench. \textbf{Left}: Representative examples illustrating the eight task categories. \textbf{Right}: Data distribution across scene categories and task types. The benchmark comprises \textbf{2,500} QA pairs over \textbf{1,181} video clips.}
\label{fig:case}
\vspace{-6mm}
\end{figure*}

\vspace{-1mm}
\section{Related Work}
\label{sec:relatedwork}

\paragraph{Spatial-aware MLLMs.}
Recent multimodal large language models (MLLMs)~\cite{an2025llava,bai2023qwen,wu2024next} integrate the reasoning capability of large language models~\cite{floridi2020gpt, bai2023qwen,touvron2023llama} with the perceptual strength of vision encoders~\cite{radford2021learning,oquab2023dinov2, he2022masked,wu2024towards}, achieving remarkable progress in visual understanding and cross-modal reasoning~\cite{yin2024survey}.
These advances have inspired growing interest in developing world models~\cite{ha2018world} and embodied agents~\cite{du2024embspatial, liu2025spatialcot}, where spatial understanding plays a central role in grounding models to the physical world.
However, achieving robust visual spatial intelligence remains a major challenge, motivating recent efforts toward constructing spatial-aware MLLMs~\cite{cheng2024spatialrgpt, wu2025spatialmllm,ouyang2025spacer,ray2024sat, huang2025mllms, zhan2025actial}.
In contrast to prior studies primarily focusing on spatial reasoning within observed scenes, our work evaluates models' spatial causal intelligence in unseen videos, aligning more closely with the human ability to infer unobserved spatial dynamics, an essential capability for real-world embodied applications.

\vspace{-3mm}
\paragraph{Benchmarking Spatial Intelligence.}
A number of benchmarks assess spatial reasoning in multimodal systems. 
Early efforts center on 2D imagery (e.g., 3DSR-Bench~\cite{ma20253dsrbench}, OmniSpatial~\cite{jia2025omnispatial}, Spatial457~\cite{wang2025spatial457} and InternSpatial-Bench~\cite{deng2025internspatial}), while VSI-Bench~\cite{yang2025thinking}, MMSI-Bench~\cite{yang2025mmsi} and All-Angles-Bench~\cite{yeh2025seeing} probe multi-view and relational reasoning, and VLM4D~\cite{zhou2025vlm4d} extends evaluation to spatio-temporal video settings. 
However, these benchmarks largely emphasize static perception or frame-level understanding and thus overlook causal dependencies unfolding over time. 
More recent dynamic evaluations such as STI-Bench~\cite{li2025sti} and DSI-Bench~\cite{zhang2025dsi} incorporate motion, yet remain predominantly perceptual. 
In contrast, our SCP-Bench explicitly targets forward and backward spatial causal reasoning in dynamic videos, offering a principled testbed for both predictive and diagnostic spatial intelligence. 
A comprehensive comparison with representative benchmarks is provided in Appendix~\S\ref{app:bench_compare}.
\vspace{-2mm}

\begin{figure*}[!t]
  \centering
\includegraphics[width=0.99\textwidth]{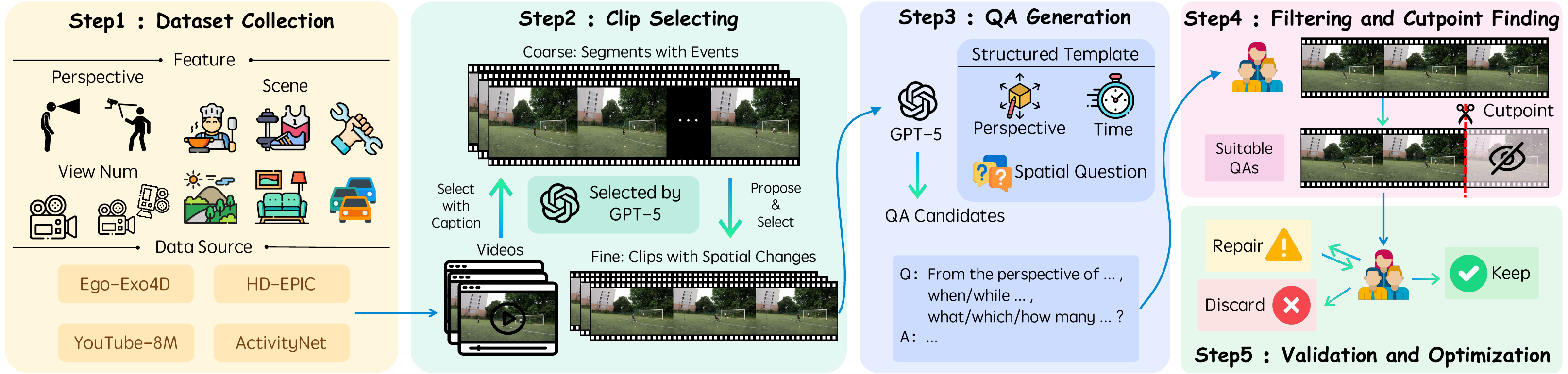}
\vspace{-2mm}
\caption{Overview of the SCP-Bench construction pipeline. The process comprises five stages: (1) collection of diverse video sources, (2) clip selection with spatially dynamic segments, (3) generation of candidate QA pairs, (4) QA filtering and cutpoint identification, and (5) dataset validation and refinement.}
\label{fig:dataset_construction}
\vspace{-6mm}
\end{figure*}
\section{SCP-Bench}
\label{sec:benchmark}
\vspace{-2mm}

\paragraph{Overview.}
We formulate SCP as a multi-choice QA task.
Formally, given a video clip with only partial temporal context, along with a question and multiple candidate options, the model aims to select the option that best reflects the underlying spatial causal reasoning.
We further introduce SCP-Bench, a benchmark designed to systematically evaluate this capability of MLLMs on the SCP task.
Following, the benchmark design is detailed in Sec.~\S\ref{sec:benchmark_design}, and the construction process is described in Sec.~\S\ref{sec:benchmark_construction}.

\vspace{-1mm}
\subsection{Benchmark Design}
\label{sec:benchmark_design}
\vspace{-2mm}
To ensure systematic and fine-grained evaluation of SCP, we design SCP-Bench around four complementary dimensions (i.e., \textit{question type}, \textit{causal direction}, \textit{perspective setting}, and \textit{scene diversity}) that together define the benchmark's structure and scope.
We detail these design components below.

\vspace{-6mm}
\paragraph{Question Type.}
We design 8 task categories to capture variations in spatial causal structure, organized by difficulty.
(1) Object-invariant identification (e.g., \textit{Relative Size}) focuses on recognizing the correct entity when object size remains constant.
(2) Attribute-dynamic reasoning (e.g., \textit{Appearance Order}, \textit{Relative Speed}, \textit{Spatial State}) probes changes in object attributes under causal evolution.
(3) Interaction-level inference (e.g., \textit{Counting}, \textit{Planning}, \textit{Relation}) requires higher-order reasoning over object interactions.
Detailed definitions are provided in  Appendix \S\ref{app:consturct}.

\vspace{-6mm}
\paragraph{Causal Direction.}
Since the model observes only the visible part before or after the cut point, the task naturally divides into two causal directions: \textit{backward} inference, which reconstructs prior states, and \textit{forward} prediction, which anticipates subsequent scene evolution.

\vspace{-6mm}
\paragraph{Perspective Setting.}
Since spatial reasoning depends on how a scene is perceived, SCP-Bench includes both single-view and multi-view settings.
The single-view setting provides an \textit{ego} (first-person) and \textit{exo} (third-person) perspective.
The multi-view setting pairs a video with a reference image from another viewpoint.
Specifically, this setting includes \textit{ego–exo} indicating that the video is observed from an ego view but answered from an exo view, with \textit{exo–ego} and \textit{exo–exo} defined analogously.

\vspace{-5mm}
\paragraph{Scene Diversity.}
To capture the breadth of environments in which spatial causal reasoning occurs, SCP-Bench includes \textit{Artistic Performances}, \textit{Animal Related}, \textit{Sports Related}, \textit{Life Records}, \textit{Factory/Machine Related}, and \textit{Driving-view Related}, ensuring that models are evaluated under diverse visual dynamics and interaction patterns.

\begin{table*}[!t]
\centering
\caption{
Evaluation on the SCP-Bench. 
``Avg.'' indicates the overall average accuracy. For each category, the best-performing
closed model and open-source model in average score are both indicated in \colorbox{deepblue}{deep blue}, and best performance on each task is \boxed{boxed}.
}
\label{tab:main_results}
\vspace{-3mm}
\rowcolors{2}{gray!8}{white}
\resizebox{\textwidth}{!}{
\begin{tabular}{lcccccccccc}
\toprule
\textbf{Model} & \cellcolor{lightblue}\bf\textbf{Avg.} & \textbf{Appearance Order} & \textbf{Counting} & \textbf{Planning} & \textbf{Relation} &   \textbf{Relative Distance} & \textbf{Relative Size} & \textbf{Relative Speed} & \textbf{Spatial State} \\
\midrule

Human Performance & \cellcolor{lightblue}89.61 & 97.60 & 81.20 & 92.26 & 85.70 & 86.70 & 97.62 & 91.61 & 84.17 \\

\hdashline
\rowcolor{lightgrey} \multicolumn{11}{c}{$\bullet$ \textit{Closed Models}} \\
GPT-5 &  \cellcolor{deepblue}\bf66.24 &  \boxed{\textbf{79.04}} &  \boxed{\textbf{58.12}} &  \boxed{\textbf{59.06}} &  \boxed{\textbf{64.07}} &  \boxed{\textbf{70.48}} &  \boxed{\textbf{95.24}} &  \boxed{\textbf{77.42}} &  \boxed{\textbf{65.11}} \\
Gemini-2.5-Pro & \cellcolor{lightblue}55.84 & 69.28 & 54.87 & 52.76 & 46.20 & 63.47 & 88.10 & 67.10 & 62.41 \\
Gemini-2.5-Flash & \cellcolor{lightblue}52.10 & 59.28 & 52.14 & 51.74 & 43.14 & 57.75 & 88.10 & 66.45 & 55.60 \\
Claude-Sonnet-4.5 & \cellcolor{lightblue}56.14 & 68.86 & 52.14 & 57.43 & 45.65 & 60.90 & 80.95 & 68.39 & 63.90 \\
\hdashline
\rowcolor{lightgrey} \multicolumn{11}{c}{$\bullet$ \textit{Open-source Models}} \\
Qwen3-VL-2B & \cellcolor{lightblue}43.04 & 41.92 & 42.74 & 45.01 & 40.85 & 44.41 & 59.52 & 47.10 & 40.65 \\
Qwen3-VL-8B & \cellcolor{lightblue}47.52 & 54.49 & 51.28 & 49.29 & 42.33 & 49.47 & 90.48 & 46.45 & 46.40 \\
Qwen3-VL-30B-A3B & \cellcolor{lightblue}54.16 & 65.27 & 52.14 & 54.79 & 46.22 & 56.65 & 85.71 & 66.45 & 57.19 \\
Qwen3-VL-32B & \cellcolor{lightblue}56.84 & 59.88 & 51.28 & 58.66 & 52.63 & 57.98 & 90.48 & 67.10 & 55.04 \\
Qwen3-VL-235B-A22B & \cellcolor{deepblue}\bf61.04 &  \boxed{\textbf{67.07}} & 54.70 & 60.90 & \boxed{\textbf{55.03}} & \boxed{\textbf{63.03}} &  \boxed{\textbf{97.62}} &  \boxed{\textbf{74.84}} &  \boxed{\textbf{63.31}} \\
Qwen3-Omni-30B-A3B & \cellcolor{lightblue}53.60 & 63.47 & 55.56 & 53.56 & 47.03 & 53.72 & 88.10 & 65.81 & 55.40 \\
InternVL3.5-8B & \cellcolor{lightblue}50.52 & 59.88 & 54.70 & 54.79 & 43.82 & 54.52 & 61.90 & 58.71 & 44.96 \\
InternVL3.5-38B & \cellcolor{lightblue}53.56 & 62.28 & 53.85 & 56.01 & 46.34 & 57.98 & 90.48 & 65.81 & 48.20 \\
InternVL3.5-241B-A28B & \cellcolor{lightblue}\bf56.96 &  \boxed{\textbf{67.07}} &  \boxed{\textbf{60.68}} &  \boxed{\textbf{61.10}} & 46.11 & 60.37 & 90.48 & 68.39 & 60.07 \\
MiniCPM-V-4.5 & \cellcolor{lightblue}43.80 & 53.29 & 49.57 & 43.99 & 36.04 & 49.20 & 76.19 & 52.26 & 42.81 \\
DeepSeek-VL2 & \cellcolor{lightblue}38.08 & 45.51 & 38.46 & 39.51 & 29.41 & 45.74 & 73.81 & 53.55 & 33.81 \\
NVILA-8B & \cellcolor{lightblue}34.40 & 36.53 & 36.75 & 38.09 & 30.66 & 30.05 & 59.52 & 38.71 & 37.05 \\
NVILA-15B & \cellcolor{lightblue}45.28 & 54.49 & 45.30 & 48.07 & 35.35 & 52.13 & 73.81 & 50.97 & 49.28 \\
LLaVA-OneVision-7B & \cellcolor{lightblue}36.48 & 42.51 & 37.61 & 37.07 & 31.24 & 38.30 & 64.29 & 46.45 & 35.61 \\
LLaVA-OneVision-70B & \cellcolor{lightblue}50.84 & 64.67 & 52.99 & 48.68 & 44.39 & 53.46 & 78.57 & 61.94 & 51.80 \\
LLaVA-OneVision-1.5-8B & \cellcolor{lightblue}45.52 & 56.29 & 47.01 & 46.44 & 39.13 & 50.27 & 80.95 & 51.61 & 41.73 \\
LLaVA-NeXT-Video-7B & \cellcolor{lightblue}36.60 & 43.11 & 25.64 & 35.44 & 29.52 & 48.40 & 54.76 & 54.84 & 32.73 \\

\hdashline
\rowcolor{lightgrey} \multicolumn{11}{c}{$\bullet$ \textit{Spatial Models}} \\
Spatial-MLLM & \cellcolor{lightblue}39.76 & 45.51 & 28.21 & 33.81 & 38.33 & 49.73 & 66.67 & 50.97 & 32.37 \\
SpaceR & \cellcolor{lightblue}41.36 & 52.10 & 34.19 & 40.53 & 34.90 & 45.21 & 59.52 & 54.19 & 44.60 \\

\bottomrule
\end{tabular}
}
\vspace{-2mm}
\end{table*}

\vspace{-1.0mm}
\subsection{Construction Process}
\label{sec:benchmark_construction}
\vspace{-1.0mm}
We show the detailed construction pipeline in Fig.~\ref{fig:dataset_construction}, which comprises the following five key steps:

\noindent$\blacktriangleright$ \textbf{Step-1: Video Source Selection.}
To ensure the various attributes of SCP-Bench, we choose the synchronized ego–exo multi-view dataset Ego-Exo4D~\cite{grauman2024ego}, the high-resolution egocentric dataset HD-EPIC~\cite{perrett2025hd}, and YouTube-based datasets YouTube-8M~\cite{abu2016youtube} and ActivityNet~\cite{caba2015activitynet}.

\noindent$\blacktriangleright$ \textbf{Step-2: Automatic Clip Screening.}
Before generating QAs, we apply an automated step to extract high-quality short clips. For each video, GPT-5~\cite{openai2025gpt5} expands the provided segments, produces dense temporally aligned captions, and filters for portions that exhibit clear spatial change. It then proposes candidate clips with explicit start–end timestamps, from which we retain the validated ones for downstream QA generation.

\noindent$\blacktriangleright$ \textbf{Step-3: Candidate QA Generation.}
To generate challenging and well-controlled QAs, we design a structured prompt template that explicitly encodes three complementary dimensions. The \textbf{perspective} fixes the viewpoint from which the question and answer should be interpreted. The \textbf{time} dimension specifies the unseen moment or temporal window, outside the visible part, that the model must infer. The \textbf{spatial question type} determines which category of spatial reasoning the QA should target. GPT-5 then uses this template to produce diverse and well-structured QA candidates for each selected clip.

\noindent$\blacktriangleright$ \textbf{Step-4: Filtering and Cutpoint Finding.}
As our task is highly sensitive to space and time, all QA candidates are manually reviewed. We retain only questions that are unambiguous, answerable from the clip, and genuinely grounded in spatial reasoning, while discarding duplicates or weakly supported items. For each approved QA, annotators first determine a precise cutpoint that separates the clip into visible and invisible parts, after which GPT-5 generates the corresponding distractor options.

\noindent$\blacktriangleright$ \textbf{Step-5: Validation and Optimization.}
To ensure QA quality, annotators conduct multiple rounds of review to refine and correct candidate items. Unanswerable or invalid samples are removed, while fixable issues, such as mismatched question types, unclear perspectives, ill-posed wording or targets, misaligned temporal windows or cutpoints, and defective options, are revised and rechecked.
After these, the remaining QAs constitute SCP-Bench.

\vspace{-2mm}
\section{Setups of Experiments and Analyses}

\vspace{-1mm}
In the following sections, we fundamentally explore the three tiers of critical questions as raised in Sec. \S\ref{sec:intro}.
To comprehensively evaluate existing models, we consider a wide range of MLLMs on SCP-Bench, including proprietary, open-source, and spatially specialized models.
Specifically, we assess four proprietary models: GPT-5~\cite{openai2025gpt5}, Gemini-2.5-Pro~\cite{comanici2025gemini}, Gemini-2.5-Flash~\cite{comanici2025gemini}, and Claude-Sonnet-4.5~\cite{anthropic2025claudeSonnet4_5}.
For open-source MLLMs, we include Qwen3-VL~\cite{yang2025qwen3}, Qwen3-Omni~\cite{xu2025qwen3}, InternVL3.5~\cite{wang2025internvl3}, MiniCPM-V-4.5~\cite{yu2025minicpm}, DeepSeek-VL2~\cite{wu2024deepseek}, NVILA~\cite{liu2025nvila}, LLaVA-OneVision~\cite{li2024llava,an2025llava}, and LLaVA-NeXT-Video~\cite{zhang2024video}.
Additionally, we evaluate two models specifically designed for spatial reasoning: Spatial-MLLM~\cite{wu2025spatialmllm} and SpaceR~\cite{ouyang2025spacer}.
All models are tested using their officially recommended configurations to ensure fair and representative results. Each model answers multiple-choice questions based on the input video, question, and answer options, with the accuracy rate serving as the primary evaluation metric.
Further implementation details are provided in Appendix~\S\ref{app:exp_detail}.

\vspace{-2mm}

\section{How Well Do Current Models Perform?}
\label{How Well Do Current Models Perform}
\vspace{-1mm}

\begin{figure}[!t]
  \centering
\includegraphics[width=0.99\columnwidth]{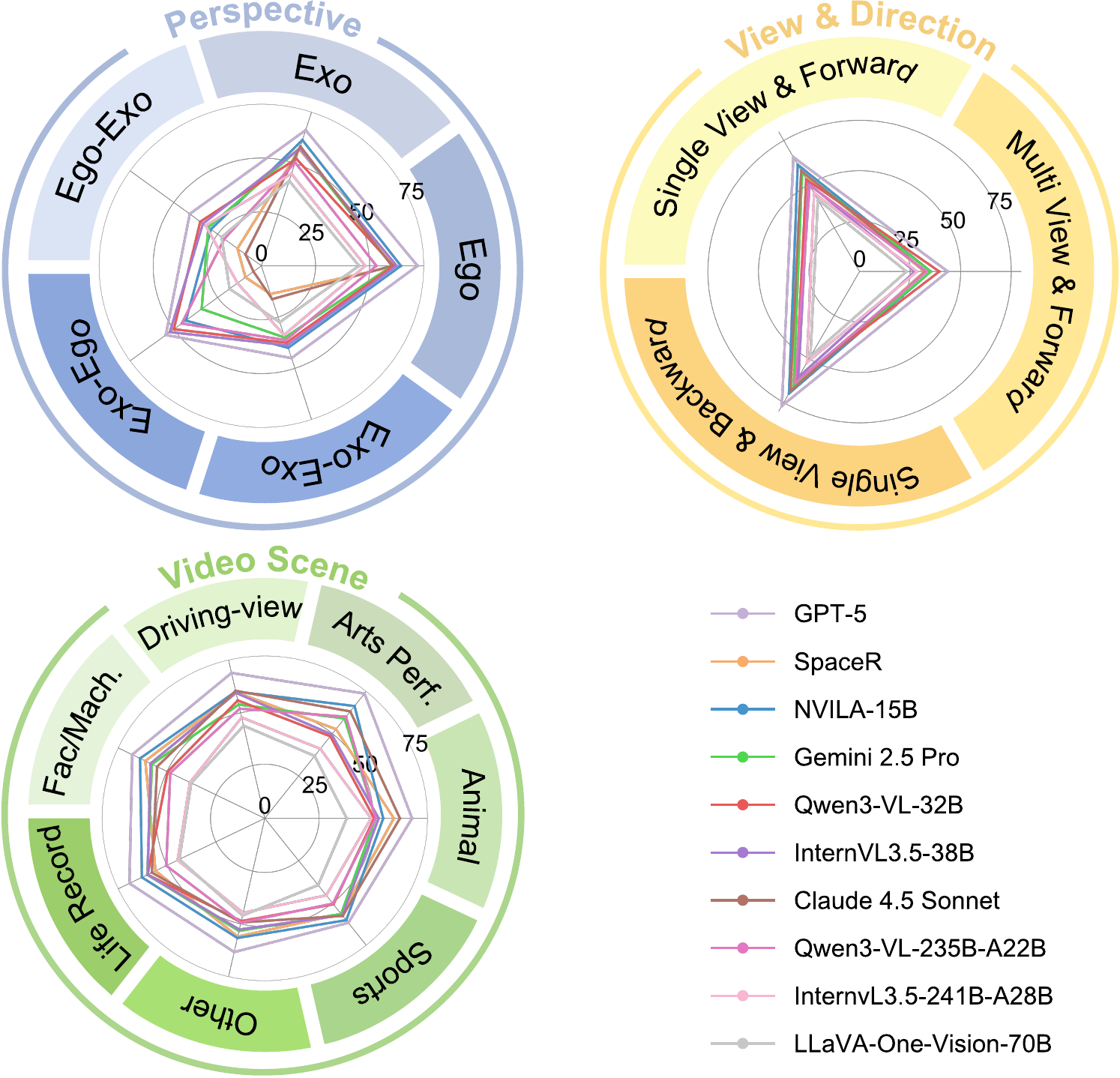}
\vspace{-2mm}
\caption{Results across perspectives, view directions, and scenes.}
\label{fig:extend_main_results}
\vspace{-7mm}
\end{figure}

\vspace{-1mm}
\paragraph{Overall Evaluation Results.}
Table~\ref{tab:main_results} summarizes the overall performance of MLLMs on SCP-Bench. Current systems remain far below human level, underscoring the substantial gap in spatial causal prediction.
GPT-5 attains the highest accuracy (66.24\%), followed by Qwen3-VL-235B (61.04\%) and InternVL3.5-241B (56.96\%).
Notably, several open-source models match or surpass proprietary ones on specific tasks, for example, outperforming GPT-5 in \textit{Counting} and \textit{Planning} and achieving comparable results in \textit{Relative Size}, \textit{Relative Speed}, and \textit{Spatial State}.
At the level of question types, the difficulty landscape becomes clearer.
\textit{Relative Size} is consistently the easiest, whereas \textit{Object Relations}, \textit{Planning}, and \textit{Counting} are the most challenging, 
as they require more abstract spatial causal reasoning and higher-order object interaction understanding.

We further examine performance across perspectives, causal directions, and scenarios (Fig.~\ref{fig:extend_main_results}). 
Models exhibit clear difficulty with multi-view prediction compared to single-view reasoning, indicating limited perspective correspondence. 
In causal directionality, models perform better when inferring past (backward) events than future (forward) ones, likely because reasoning from known outcomes is easier than anticipating unseen consequences.
Finally, model performance remains relatively balanced across different scene categories, with slightly stronger results in driving-related and factory/machine environments.

\begin{figure}[!t]
    \centering
    \includegraphics[width=0.80\linewidth]{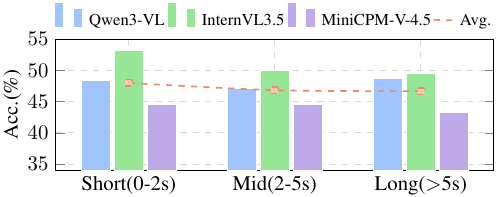}
    \vspace{-2mm}
    \caption{Temporal extrapolation horizon analysis. Samples are grouped by the time gap between the cutpoint and future event: short (0–2s), mid (2–5s), and long ($>$ 5s). 
    }
    \label{fig:time_range}
    \vspace{-6mm}
\end{figure}

\vspace{-5mm}
\paragraph{The Impact of Temporal Extrapolation Horizon.}
We analyze how model performance varies with different temporal prediction ranges. Specifically, we divide the future duration between the cutpoint and the event completion into three intervals: short (0–2s), mid (2–5s), and long ($>$5s). We compare three models of comparable size (8B), as shown in Fig.~\ref{fig:time_range}.
Overall, model accuracy remains relatively stable across horizons, averaging around 46.8\%. 
This indicates that dynamic frame sampling in existing MLLMs mitigates sensitivity to temporal length difference, and also that the current temporal segmentation range may be too narrow to induce significant variation.

\begin{figure*}[!t]
  \centering
   \includegraphics[width=0.99\linewidth]{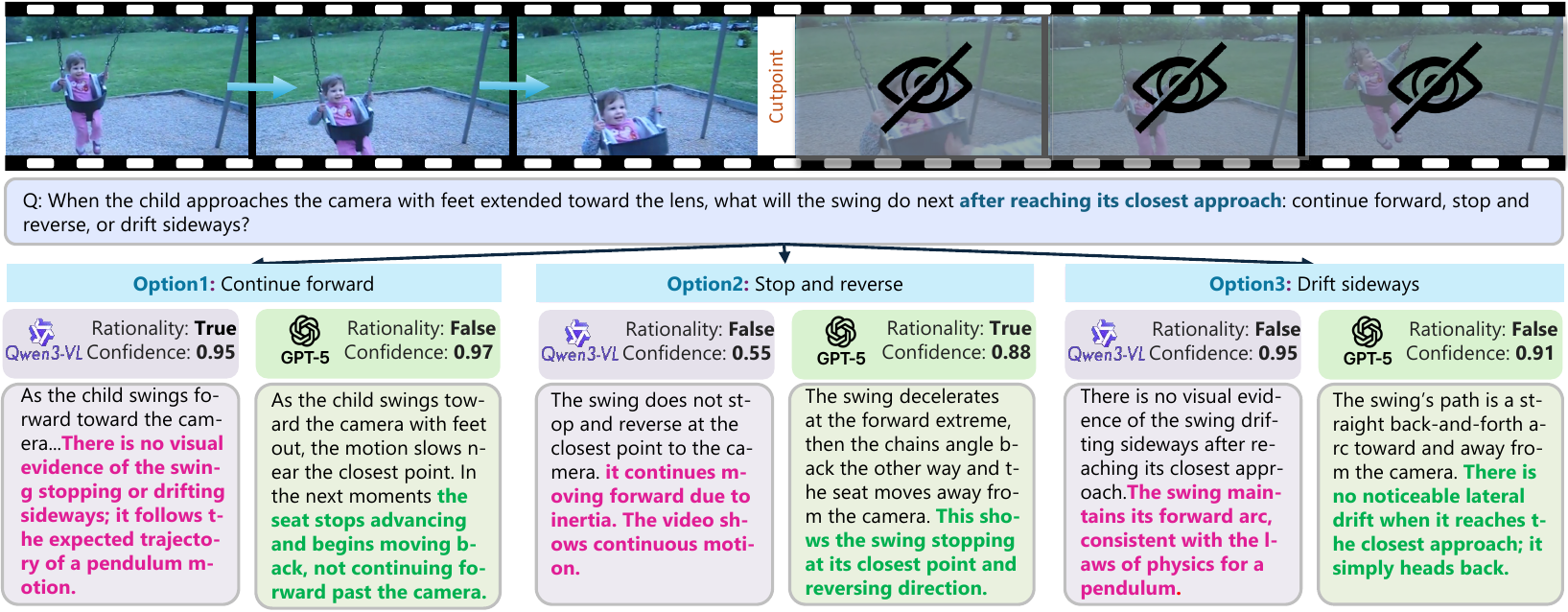}
   \vspace{-2mm}
   \caption{
   Causal consistency evaluation. Each option shows the model's reasoning rationality, confidence score, and explanation, assessing whether MLLMs can infer spatial order consistent with physical and causal constraints.
   }
   \label{fig:Causal_case}
   \vspace{-4mm}
\end{figure*}

\vspace{-5mm}
\paragraph{Causal Consistency Evaluation.}
To complement the above quantitative results, we conduct a case study to examine whether models obey basic physical and temporal causal constraints. 
Given a video of a child on a swing, we construct three candidate event progressions after the swing reaches its closest point to the camera: continue forward, stop and reverse, or drift sideways. 
For each option, the model outputs a rationality judgment (Rationality), a confidence score (Confidence), and an explanation grounded in visible evidence frames under a fixed reference frame. 
As illustrated in Fig.~\ref{fig:Causal_case}, Qwen3-VL assigns high rationality and confidence to the ``continue forward'' option, emphasizing local motion continuity while ignoring the physical constraint that the swing should reverse at the turning point. 
In contrast, GPT-5 explicitly refers to the deceleration, brief stop, and backward motion near the extremum, and correctly rejects this anti-causal trajectory. 
This case shows that even when models perceive local motion correctly, their explanations can still violate basic causal constraints, revealing a gap in causal consistency for MLLMs.

\begin{table}[!t]
    \centering
    \fontsize{9}{11}\selectfont
    \setlength{\tabcolsep}{1.05mm}
    \begin{tabular}{lccc}
    \toprule
    Setting     & Qwen3-VL &  InternVL3.5 & MiniCPM-V-4.5 \\
    \midrule
    Base & 47.52 & 50.52 & 43.80 \\
    \hdashline
    Gold Video & 54.96$_{\textcolor{red}{\uparrow 7.44}}$ & 58.64$_{\textcolor{red}{\uparrow 8.12}}$ & 43.72$_{\textcolor{mygreen}{\downarrow 0.08}}$\\
    Caption w/o Video & 46.76$_{\textcolor{mygreen}{\downarrow 0.76}}$& 41.24$_{\textcolor{mygreen}{\downarrow 9.28}}$ & 45.76$_{\textcolor{red}{\uparrow 1.96}}$\\
    \bottomrule
    \end{tabular}
    \vspace{-2mm}
    \caption{
    Comparison between perception and reasoning. ``Base'' is the standard evaluation performance; ``Gold Video'' evaluates perceptual understanding using unseen part; ``Caption w/o Video'' tests reasoning based on captions alone. \textcolor{red}{Red} and \textcolor{mygreen}{green} indicate performance gains and drops, respectively.
    }
    \label{tab:perception_reasoning}
    \vspace{-2mm}
\end{table}

\begin{figure}[!t]
  \centering
  \vspace{-2mm}
\includegraphics[width=0.90\columnwidth]{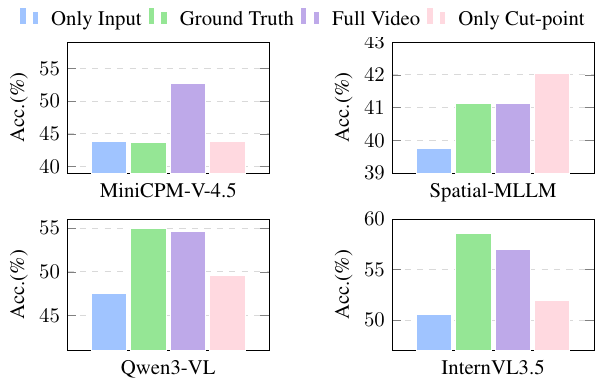 }
\vspace{-4mm}
\caption{Visible Range Comparison. \textit{Only cutpoint} uses the cutpoint frame as input; \textit{Full video} provides the entire clip; \textit{Ground truth} includes only the unseen clips adjacent to the cutpoint.
}
\label{fig:visible_range}
\vspace{-7mm}
\end{figure}

\begin{figure}
    \centering
    \includegraphics[width=0.95\linewidth]{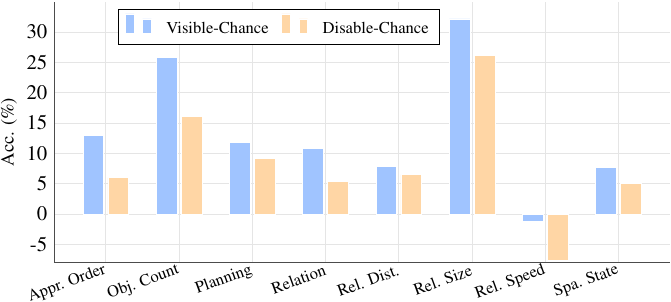}
    \vspace{-2mm}
    \caption{Performance gap comparison. Accuracy improvements over the chance level for vision-enabled (w/ video) and vision-disabled (w/o video) settings across task categories.}
    \label{fig:visual_no_visual}
    \vspace{-2mm}
\end{figure}

\vspace{-1mm}
\begin{takeawaybox}[Performance in Spatial Causal Prediction]
\begin{compactitem}
    \item \textit{SCP is a formidable challenge for current MLLMs.}
    \item \textit{Large open-source models perform on par with closed ones.}
    \item \textit{MLLMs perform similarly across time ranges.}
\end{compactitem}
\end{takeawaybox}

\section{What Affects Spatial Causal Prediction?}
\label{sec:affects}
\vspace{-1mm}
\paragraph{Perception vs. Reasoning Decomposition.}
A robust solution to spatial causal prediction requires two essential components: accurate perception of the visual input and reliable reasoning built upon that perception. 
To investigate which factor plays a more critical role in their poor performance, we design a controlled set of experiments. 
In one condition, we provide the models with the unseen parts of the clips, referred to as the Gold Video, thereby removing the need for causal inference and turning the task into pure visual understanding.
In contrast, we replace the visible parts of the clips with dense captions generated by Tarsier~\cite{wang2024tarsier}, thereby isolating perception and forcing the model to rely solely on textual reasoning.
Table~\ref{tab:perception_reasoning} shows that the model achieves an average accuracy of 52.44\% with the Gold Video input, compared to 44.59\% with dense captions, indicating that perception is not the primary bottleneck and that the main limitation lies in spatial causal reasoning.

\vspace{-4mm}
\paragraph{Single-Frame vs. Multi-Frame Reasoning.}
To examine whether models truly leverage temporal continuity rather than static cues, we compare single-frame and multi-frame reasoning. Specifically, we evaluate model performance when given only the cutpoint frame versus the visible part. As shown in Fig.~\ref{fig:visible_range}, all four models show a slight improvement over the multi-frame condition when performing single-frame reasoning. 
This counterintuitive result suggests that temporal cues contribute minimally to model performance under the base spatial causal reasoning task setting, and the observed gains likely stem from static spatial perception rather than genuine temporal understanding.

\begin{table}[!t]
  \centering
      \fontsize{9}{11}\selectfont
\setlength{\tabcolsep}{0.5mm}
  \begin{tabular}{@{}lcccccc}
    \toprule
    Setting & Qwen3-VL & InternVL3.5 & MiniCPM-V-4.5 & LLaVA-OV-1.5\\
    \midrule
    Base & 47.52 & 50.52 & 43.80  & 45.52 \\
    \hdashline
    w/ Flip & 46.76$_{\textcolor{mygreen}{\downarrow 0.76}}$ & 48.56$_{\textcolor{mygreen}{\downarrow 1.96}} $ &  42.24$_{\textcolor{mygreen}{\downarrow 1.56}} $ & 45.32$_{\textcolor{mygreen}{\downarrow 0.2}} $ \\ 
    w/ CoT & 51.52$_{\textcolor{red}{\uparrow 4.0}} $ & 50.36$_{\textcolor{mygreen}{\downarrow 0.16}} $ & 44.88$_{\textcolor{red}{\uparrow1.08}} $ & 42.88$_{\textcolor{mygreen}{\downarrow2.64}} $ \\
    \bottomrule
  \end{tabular}
  \vspace{-2mm}
    \caption{Comparison with flipped video input and CoT reasoning. ``w/ Flip'' reverses the input video temporally; ``w/ CoT'' applies step-by-step reasoning.}
  \label{tab:flip_cot}
  \vspace{-3mm}
\end{table}

\vspace{-4mm}
\paragraph{Vision Causal Perception.}
To further evaluate whether MLLMs can perceive spatial causal logic in the video input, we temporally reverse each input clip while keeping the question unchanged and compare performance with the original forward videos. As shown in Table~\ref{tab:flip_cot}, all models show a slight accuracy drop after reversal, with the largest drop being 1.96\%. This small drop indicates limited sensitivity to temporal inversion, suggesting that current models have yet to develop a robust notion of spatial causality.
\vspace{-3mm}

\vspace{-1mm}
\paragraph{Visible Range Comparison.}
We investigate how the visible range affects model performance. 
As shown in Fig.~\ref{fig:visible_range}, models perform best when given the Gold Video input, where direct access to the answer period eliminates the need for spatial causal reasoning and allows performance gains through pure visual understanding. 
The full-video setting yields slightly lower accuracy but remains above the base condition, suggesting that when the Gold Video part is seen, the visible part introduces mild noise that weakens comprehension. 
These results indicate that MLLMs perform well when the demand for SCP is minimal, but struggle to infer spatial causality beyond direct observation.

\vspace{-3mm}
\paragraph{Is Visual Information Necessary?}
As the textual capabilities of models continue to improve, models can sometimes derive correct answers from pure textual reasoning based solely on the knowledge gained during training. This raises the question: can spatial causal reasoning be achieved through pure text alone? We compare model performance with and without video input. 
The experimental results in Fig.~\ref{fig:perception_enhancement} show that the model's accuracy drops significantly when no video is provided. 
Furthermore, we used the Tarsier to generate dense captions for the video; however, even when no video is provided but an auxiliary dense caption is available, the model's performance still declines. 
This suggests that although dense captions can partially compensate for the absence of visual input, visual information itself remains indispensable for spatial causal prediction.

\begin{takeawaybox}[What Affects Spatial Causal Prediction]
\begin{compactitem}
    \item \textit{Perception and Reasoning capabilities limit SCP.}
    \item \textit{Visual information is important for SCP.}
    \item \textit{MLLMs cannot construct temporal logic to complete reasoning within a video.}    
\end{compactitem}

\end{takeawaybox}

\section{How to Improve Spatial Causal Prediction?}
\label{sec:Improvement}

\begin{figure}[!t]
  \centering
  \includegraphics[width=0.95\linewidth]{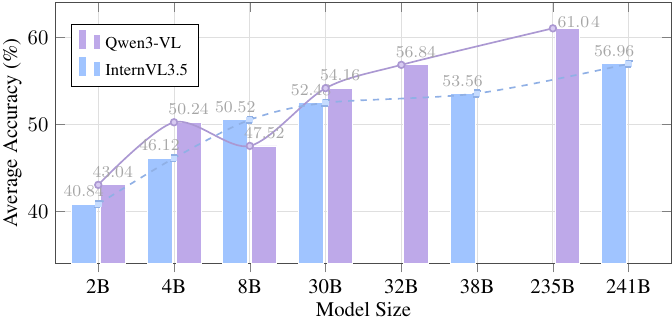}
  \vspace{-2mm}
    \caption{Performance comparison on different model sizes.}
    \label{fig:model_size}
    \vspace{-4mm}
\end{figure}

\vspace{-2mm}
\paragraph{Model Scale-up Effect Analysis.}
We investigate whether enlarging model size leads to improved performance on SCP-Bench by comparing different variants of Qwen3-VL and InternVL3.5 across scales, as depicted in Fig.~\ref{fig:model_size}.
Overall, performance exhibits a clear upward trend with increased model size, confirming the positive correlation between scaling and spatial reasoning capability.
However, the improvement is not strictly monotonic, i.e., small-scale models (e.g., 4B vs. 8B) sometimes show performance fluctuations, suggesting that scaling benefits become more stable and pronounced only when the model grows by a substantial order of magnitude (e.g., 4B $\rightarrow$ 30B).

\vspace{-4mm}
\paragraph{CoT Reasoning.}
We further explore the effect of incorporating a vanilla CoT prompt~\cite{wei2022cot} by adding the phrase ``think step by step'' in the prompt during inference. 
As shown in Table~\ref{tab:flip_cot}, this simple strategy yields modest improvements for certain models, for example, Qwen3-VL gains over 4\% in accuracy.
However, the effect is not consistent across architectures; some models show minimal sensitivity or even slight performance degradation, suggesting that the effectiveness of CoT prompting varies with model design and reasoning alignment.

\begin{figure}[!t]
  \centering
  \includegraphics[width=0.90\linewidth]{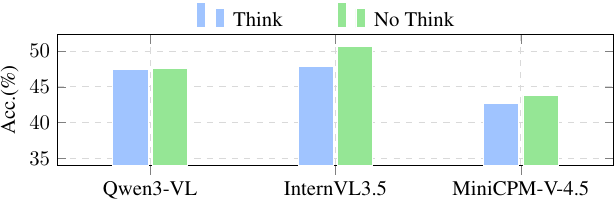}
  \vspace{-2mm}
    \caption{Performance with vs. without self-think reasoning.}
    \label{fig:think_mode}
    \vspace{-2mm}
\end{figure}

\vspace{-4mm}
\paragraph{Self-Think Reasoning.}
Recent MLLMs increasingly integrate self-think reasoning through reinforcement learning. 
To assess the generalizable effectiveness in spatial causal prediction, we compare their performances with and without think-mode reasoning. 
As shown in Fig.~\ref{fig:think_mode}, enabling think-mode does not yield consistent improvements; in fact, most models exhibit slight performance degradation, likely due to overextended reasoning chains that introduce noise and divert attention from essential spatial cues.

\begin{figure}[!t]
    \centering
    \includegraphics[width=0.99\linewidth]{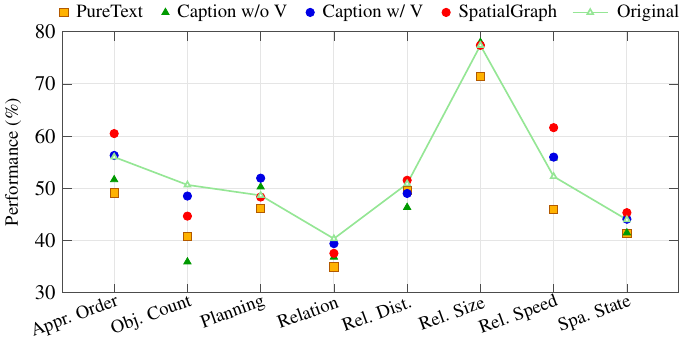}
    \vspace{-2mm}
    \caption{Comparison of perception enhancement strategies. \texttt{PureText} uses only the question; \texttt{Caption w/ V} combines dense video captions with the video and question; \texttt{Caption w/o V} uses only captions with the question; \texttt{SpatialGraph} introduces spatial-interaction graph; \texttt{Original} is the baseline.}
    \label{fig:perception_enhancement}
    \vspace{-5mm}
\end{figure}

\vspace{-4mm}
\paragraph{Perception Enhancement Strategy.}
To address the limited perceptual capability of MLLMs, we explore several mechanisms designed to enhance spatial perception: (1) generating dense captions of the input video clip to enrich scene perception, and (2) constructing spatial-interaction graphs via prompts that capture key objects, environmental elements, and their spatial and interaction relations. 
The results in Fig.~\ref{fig:perception_enhancement} indicate that these perception enhancement strategies lead to only marginal improvements overall, with noticeable gains primarily in specific tasks such as \textit{Appearance Order} and \textit{Relative Speed}. 
The models fail to leverage spatial-interaction graphs for accurate spatial causal reasoning.
When using dense captions to enhance input, the models also exhibit limited benefit, suggesting these perception-level augmentations alone are insufficient to substantially boost spatial causal reasoning.

\begin{figure}[!t]
    \centering
    \includegraphics[width=0.99\linewidth]{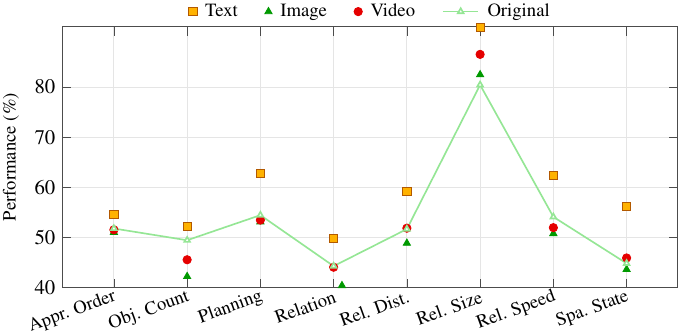}
    \vspace{-2mm}
    \caption{Unseen spatial causal scaffolds for enhanced reasoning. \texttt{Text} provides textual descriptions of future spatial states; \texttt{Image} employs generated future spatial images; \texttt{Video} uses generated future causal videos.}
    \label{fig:future_enhancement}
    \vspace{-6mm}
\end{figure}

\vspace{-4mm}
\paragraph{Causal Prediction Enhancement Strategy.}
We further investigate which types of unseen spatial causal scaffolds are able to effectively enhance reasoning performance. 
Specifically, we evaluate three forms of auxiliary information: textual descriptions generated by GPT-5~\cite{openai2025gpt5}, the future spatial images produced by FLUX.1-dev-12B~\cite{flux2024}, and the future causal videos generated by Wan2.2-TI2V-5B~\cite{wan2025wan}.
As shown in Fig.~\ref{fig:future_enhancement}, incorporating textual future predictions consistently improves performance across all tasks, likely because MLLMs are inherently more adept at processing and reasoning over textual information.
In contrast, image- and video-based scaffolds provide limited gains, likely due to input length constraints, modality noise, and the inherent perception limitations of current MLLMs. 
Nevertheless, videos outperform images in dynamic-related tasks (e.g., Relative Size and Spatial State), benefiting from their richer temporal cues.

\begin{takeawaybox}[How to Improve Spatial Causal Prediction]
\begin{compactitem}
    \item \textit{Notably increasing model size helps with SCP.}
    \item \textit{Unseen spatial causal scaffolds can effectively enhance model performance.}
    \item \textit{Vanilla CoT and self-thinking mechanisms lead to limited improvements.}
\end{compactitem}
\end{takeawaybox}

\section{Conclusion}
\label{sec:conclusion}
\vspace{-1mm}
We introduce Spatial Causal Prediction (SCP) and SCP-Bench, establishing a new paradigm for predictive spatial reasoning beyond visible scenes. 
Extensive evaluations indicate that current MLLMs remain far from human-level performance, performing better on past inference than future prediction, and relying mainly on perceptual cues. 
In-depth controlled analyses show that reasoning, rather than perception, constitutes the major bottleneck. 
While explicit reasoning and structured spatial representations bring limited gains, notably scaling up and integrating causal scaffolds offer a promising path for better SCP performance.

\section*{Acknowledgement}
This work is supported by the Ministry of Education, Singapore, under its MOE AcRF TIER 3 Grant (MOE-MOET32022-0001).

{
    \small
    \bibliographystyle{ieeenat_fullname}
    \bibliography{main}
}

\clearpage
\appendix

\makeatletter
\let\addcontentsline\origaddcontentsline
\makeatother
\input{suppl}

\end{document}

%% file: preamble.tex
\usepackage{CJKutf8}

\usepackage{CJKutf8}

\usepackage[table]{xcolor}
\usepackage{booktabs,multirow,array}
\usepackage{array}
\usepackage{makecell}
\usepackage{tikz}
\usepackage{arydshln} 
\usepackage{colortbl}
\usepackage{amsthm}
\usepackage{amsmath, amssymb}
\usepackage{adjustbox}
\usepackage{changepage}
\usepackage{tabularx}
\usepackage{array} 
\usepackage{lipsum}
\usepackage{paralist}
\usepackage{makecell}
\usepackage{pifont}
\usepackage{comment}
\usepackage{graphicx}
\usepackage{xcolor}
\usepackage{pifont}
\usepackage{bbding}
\usepackage{fontawesome}
\usepackage{xspace}
\usepackage{float}
\usepackage{algorithm}
\usepackage{listings}
\usepackage{algcompatible}
\usepackage{algpseudocode}
\usepackage{soul}
\usepackage{array}
\usepackage{makecell}
\usepackage{rotating}
\usepackage{enumitem}
\usepackage{CJKutf8}
\usepackage[most]{tcolorbox}

\definecolor{cvprblue}{rgb}{0.21,0.49,0.74}
\usepackage[pagebackref,breaklinks,colorlinks,allcolors=cvprblue]{hyperref}

\definecolor{lightblue}{RGB}{232,240,255}
\definecolor{deepblue}{RGB}{160,190,255}
\definecolor{lightgrey}{RGB}{234,234,234}
\definecolor{lightyellow}{RGB}{255,251,216}
\definecolor{mygreen}{RGB}{31,229,113}

\definecolor{takeawayblue}{RGB}{230,243,255}
\definecolor{titleblue}{RGB}{78,146,222}
\definecolor{infoblue}{RGB}{128,135,255}
\definecolor{infodeepblue}{RGB}{219,222,255}

\newtcolorbox{takeawaybox}[1][]{
  colback=takeawayblue!60,        
  colframe=titleblue,             
  coltitle=white,                 
  fonttitle=\footnotesize,            
  enhanced,
  boxrule=0pt,                   
  frame hidden,
  borderline west={3pt}{0pt}{titleblue}, 
  sharp corners,
  breakable,                     
  title={\textcolor{white}{\faLightbulbO\ Key Takeaways: #1}},
  attach boxed title to top left={yshift=-1mm,xshift=2mm},
  boxed title style={
    colback=titleblue,
    sharp corners,
    boxrule=0pt,
    interior style={fill=titleblue,opacity=1},
    breakable,              
    text width=\linewidth-1.5cm,  
  },
  before skip=8pt,
  after skip=8pt,
  left=0mm, right=2mm, top=1mm, bottom=0mm,
}

\newtcolorbox{infobox}{
  colback=takeawayblue!60,          
  colframe=titleblue,               
  boxrule=0pt,                      
  frame hidden,                     
  borderline west={3pt}{0pt}{titleblue}, 
  enhanced,
  sharp corners,        
  left=1mm, right=2mm, top=0.5mm, bottom=0.5mm,
  before skip=6pt, after skip=3pt
}

\definecolor{qblue}{RGB}{66,133,244}
\definecolor{qback}{RGB}{245,248,252}

\newtcolorbox{qbox}{
  enhanced,
  colback=qback,
  borderline west={2pt}{0pt}{qblue},
  boxrule=0pt,
  sharp corners,
  left=3pt, right=3pt, top=2pt, bottom=2pt,
  before skip=6pt, after skip=6pt
}

\definecolor{startblue}{RGB}{85,150,255}
\definecolor{endblue}{RGB}{150,200,255}

\newtcolorbox{gradientqbox}{
  enhanced,
  colback=white,
  frame hidden,
  borderline west={3pt}{0pt}{startblue!80!endblue},
  left=4pt, right=2pt, top=2pt, bottom=2pt,
  drop shadow={shadow xshift=0.7ex, shadow yshift=-0.7ex, opacity=0.1}, 
  sharp corners,
  before skip=6pt, after skip=6pt
}

\definecolor{qblue}{RGB}{78,146,222}

\newtcolorbox{qcard}{
  enhanced,
  colback=blue!3,
  colframe=qblue,
  boxrule=0pt,
  borderline west={3pt}{0pt}{qblue},
  left=5pt, right=5pt, top=3pt, bottom=3pt,
  overlay={
    \node[anchor=north east, font=\sffamily\bfseries\small, text=qblue]
    at ([xshift=-3pt,yshift=-1pt]frame.north east) {\faQuestionCircle};
  },
}

\newtcolorbox{bannerqbox}{
  enhanced,
  colback=blue!8,
  colframe=blue!40!black,
  fontupper=\bfseries,
  borderline west={3pt}{0pt}{blue!70!black},
  attach boxed title to top left={yshift=-2mm,xshift=2mm},
  boxed title style={
    colback=blue!70!black, colupper=white,
    fontupper=\bfseries\sffamily,
    sharp corners,
  },
  title={Q1},
  left=4pt, right=4pt, top=3pt, bottom=3pt
}

\usepackage{tikz}
\newcommand{\circnum}[1]{
  \tikz[baseline=(char.base)]{
    \node[shape=circle,draw,fill=black,inner sep=1pt,
          text=white,font=\sffamily\bfseries\footnotesize](char){#1};
  }
}

%% file: suppl.tex
\definecolor{cvprblue}{rgb}{0.21,0.49,0.74}
\definecolor{my_green}{RGB}{51,102,0}
\definecolor{my_red}{RGB}{204, 0, 0}
\renewcommand{\checkmark}{\textcolor{my_green}{\ding{51}}} 
\newcommand{\crossmark}{\textcolor{my_red}{\ding{55}}}
\definecolor{lightgray}{RGB}{235,235,235}

\def\paperID{xxxx}
\def\confName{CVPR}
\def\confYear{2026}

\maketitlesupplementary

\tableofcontents

\thispagestyle{empty}
\appendix

\newpage

\section{Limitations}
\label{app:limit}
Although SCP-Bench provides a rigorous and carefully structured evaluation of spatial causal prediction, it still entails several potential limitations simply because the domain itself is inherently complex and difficult to capture in full.

\subsection{Limitations in Spatial Pattern Coverage}
SCP-Bench spans a wide range of spatial patterns, yet the space of possible configurations in real-world videos is effectively unbounded.
Some highly specialized or rare spatial structures are therefore naturally beyond the scope of the current benchmark.

\subsection{Inherent Limits of Scene Variety}
Although the dataset includes a broad collection of scenes, real-world environments exhibit far greater variability than any finite benchmark can capture.
Certain niche or atypical scenarios inevitably remain under-represented.

\subsection{Data Scale Bounded by Quality Assurance}
SCP-Bench adopts a controlled data scale, as each item must pass rigorous checks on perspective, temporal clarity, and spatial precision. 
Meeting these requirements demands substantial validation effort, which keeps the dataset compact while preserving consistency and high fidelity.

\subsection{Scope and Limitation of Causality in SCP}
SCP-Bench examines models' SCP capability through inference over unseen spatial state transitions under partial observability. Explicit causal annotations, such as interventions, counterfactuals, or causal graphs, are not yet incorporated and remain for future work.

\section{Benchmark Construction Details}
\label{app:consturct}

\subsection{Video Sources}
\label{app:video_source}
To ensure that SCP-Bench provides comprehensive and reliable evaluation coverage, we carefully considered both viewpoint diversity and scene variability when selecting video sources. 
The main data sources include:
\begin{compactitem}
    \item \textbf{Ego-Exo4D}~\cite{grauman2024ego} provides synchronized egocentric and multiple aligned exocentric viewpoints across activities such as sports, household repairs, and everyday human interactions. Its multi-view alignment is essential for constructing multi-perspective questions in SCP-Bench. 
    
    \item \textbf{HD-EPIC}~\cite{perrett2025hd} contributes high-resolution egocentric recordings focused primarily on kitchen environments. These videos offer well-controlled, densely annotated footage but exhibit limited scene diversity due to their constrained capture settings.

    \item \textbf{YouTube-8M}~\cite{abu2016youtube} supplies large-scale, in-the-wild content spanning thousands of scene categories and diverse physical contexts. Its broad coverage helps enrich SCP-Bench with varied object configurations, dynamic motions, and unconstrained environments.

    \item \textbf{ActivityNet}~\cite{caba2015activitynet} contains long-form, human-centric activity videos with rich temporal structure and complex multi-object interactions. These properties make it a valuable source for identifying segments with clear causal dynamics and predictable spatial evolution.
\end{compactitem}
From these large-scale datasets, we perform careful, multi-stage filtering to extract clips that satisfy the requirements of spatial causal prediction.
This ensures that the final benchmark maintains high-quality visual content, diverse physical scenarios, and balanced coverage across scene categories, perspectives, causal directions, and question types.

\subsection{QA Candidates Generation Details}
\label{app:QA_gen}

To construct a large pool of QA candidates, we designed a fully automated generation pipeline with a two-stage filtering procedure that narrows data from long videos to segments and then to short clips, since spatial causal reasoning questions typically concern only specific portions of a video.
QA candidates are generated around these clips to ensure precise alignment with the underlying events and their temporal boundaries.

\noindent\textbf{$\blacktriangleright$ Segments Selecting.}
Starting from event-level structure, we first collect the segment annotations provided by the original datasets and then use GPT-5 to propose additional plausible segments directly from the videos.
To further control the quality of the selected segments, GPT-5 is prompted to (i) generate a detailed description for each candidate segment and (ii) filter these candidates based on multiple criteria, including spatial dynamics, event coherence, and content clarity.
This procedure yields segments that feature well-defined events and distinct spatial changes.

\noindent\textbf{$\blacktriangleright$ Clips Selecting.}
From the refined set of segments, we next identify the specific portions that exhibit meaningful spatial evolution.
GPT-5 is prompted to propose candidate clip boundaries for each segment under the constraint that every clip must present a clear and coherent trajectory of spatial change, while avoiding rapid shot transitions, severe motion blur, and other forms of visual degradation.
To ensure reliability, each proposed clip is then independently re-evaluated by GPT-5 using only the clip itself.
Clips displaying weak spatial variation, ambiguous dynamics, or unstable visual quality are discarded.
This process yields a curated collection of high-quality clips that are well-suited for generating SCP QA pairs.

\noindent\textbf{$\blacktriangleright$ QA Generation.}
Given each selected clip, we design a prompting scheme that guides GPT-5 to generate reliable spatial causal prediction questions.
Because spatial reasoning is inherently sensitive to \textit{perspective}, every question is required to explicitly specify its viewpoint, either a camera view or the perspective of a salient object, to avoid ambiguous interpretations.
In line with our definition of spatial causal prediction, each question must also clearly identify the \textit{unseen temporal region} that the model needs to infer from the visible portion of the clip.
The \textit{spatial question component} then specifies the core phenomenon being tested and is constrained to one of our eight spatial reasoning categories.
To further guarantee answer determinism, the prompting template enforces precise spatial references, such as relative positions, motions, and object interactions, so that each question is tightly grounded in visual evidence and admits a single, verifiably correct answer.

Using this pipeline, GPT-5 produces a large set of high-quality spatial causal prediction QA candidates that serve as the foundation for SCP-Bench.

\subsection{Manual Annotation Details}
\label{app:QA_selection}
Following automatic QA generation, human annotators perform multiple rounds of selection, validation, and refinement to obtain the final version of SCP-Bench.
To support this workflow, we develop 3 annotation tools that help annotators through (i) filtering QA candidates and defining cutpoints, (ii) validating retained items, and (iii) refining correctable items during annotation.

\begin{figure*}[!t]
\centering
\includegraphics[width=0.99\linewidth]{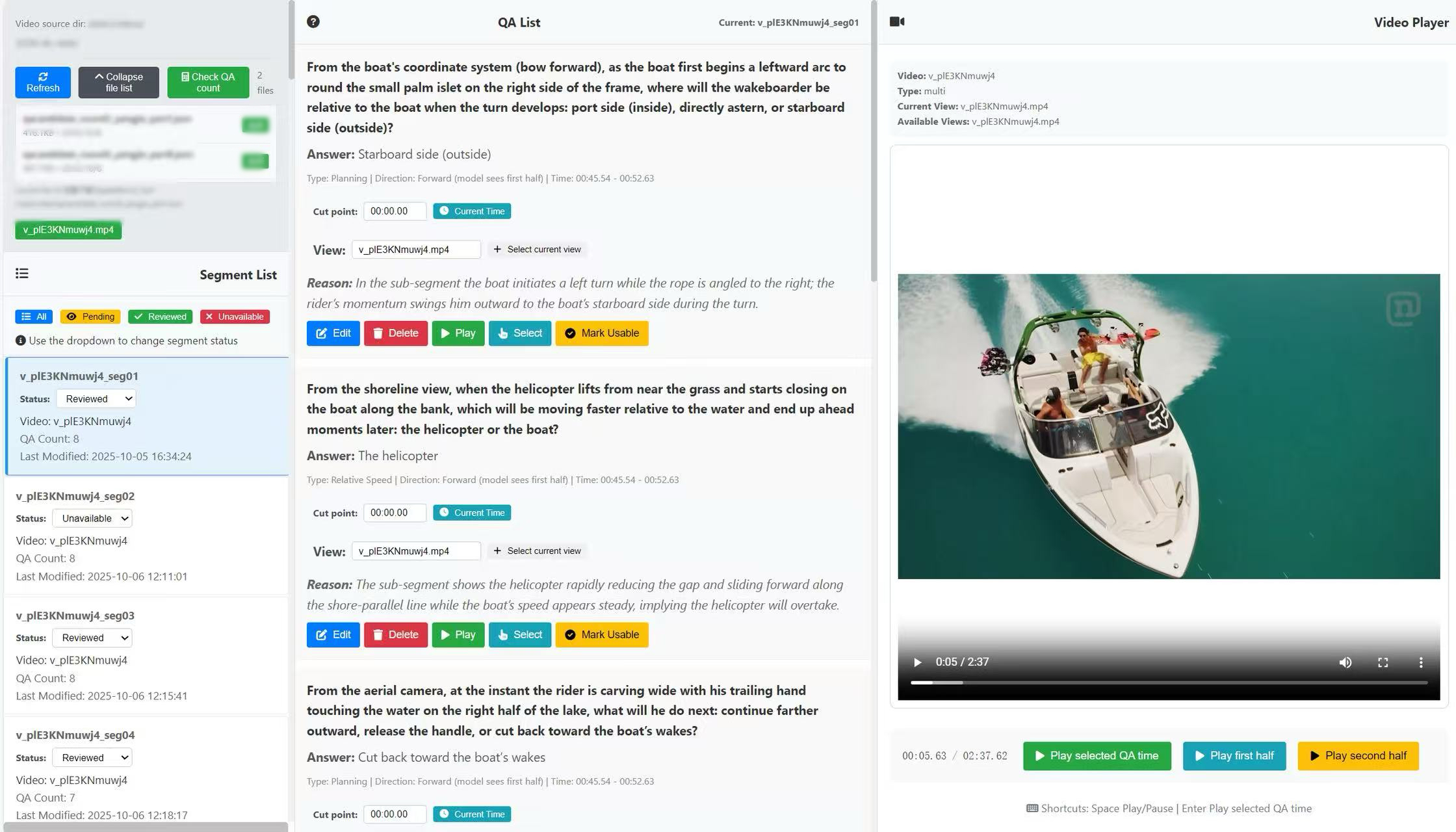}
\vspace{-2mm}
\caption{Manual filtering tool demonstration. Annotators use it to filter appropriate QA candidates and determine their cutpoints.}
\label{fig:filter_tool}
\end{figure*}

\paragraph{Filtering and Cutpoint Finding.}
Annotators first review the automatically generated candidates and determine which items satisfy the SCP task requirements.
A key step in this stage is cutpoint identification, where annotators designate the timestamp separating the visible portion of the clip from the unseen region.
These cutpoints ensure that the answer is not directly observable in the visible portion, yet remains inferable through spatial causal reasoning grounded in that visible evidence.
As illustrated in Fig.~\ref{fig:filter_tool}, the filtering tool allows annotators to inspect each clip alongside its candidate questions, decide whether a question should be retained, and assign the corresponding cutpoint.
The tool additionally provides a split-view interface, enabling annotators to preview the visible and unseen parts of the clip independently.
For each retained question, GPT-5 subsequently generates additional distractor options to complete the multiple-choice set.

\begin{figure*}[!t]
\centering
\includegraphics[width=0.99\linewidth]{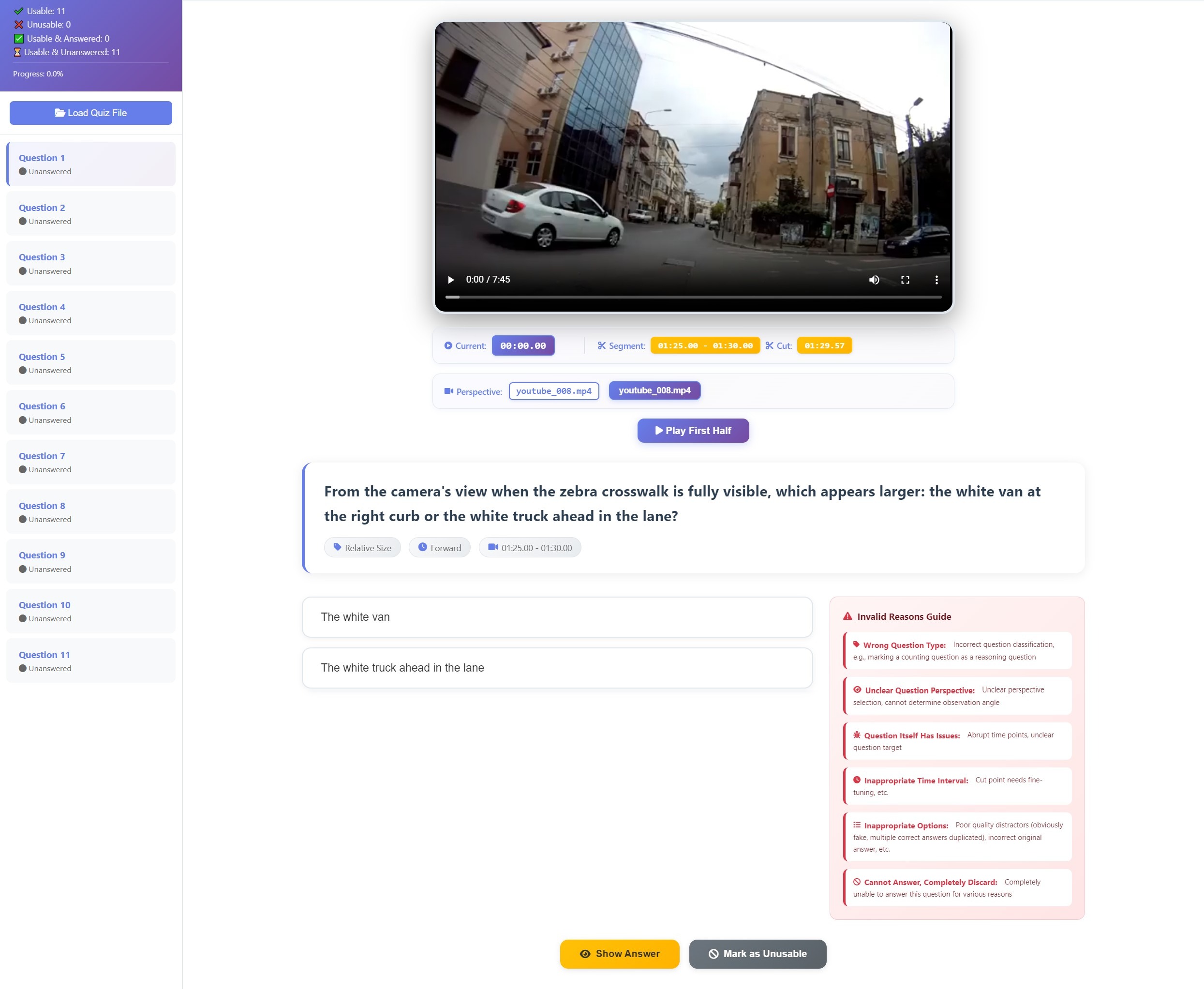}
\vspace{-2mm}
\caption{Validation tool demonstration. Annotators use it to validate each item in strict accordance with the SCP task specification.}
\label{fig:validate_tool}
\end{figure*}

\begin{figure*}[!t]
\centering
\includegraphics[width=0.9\linewidth]{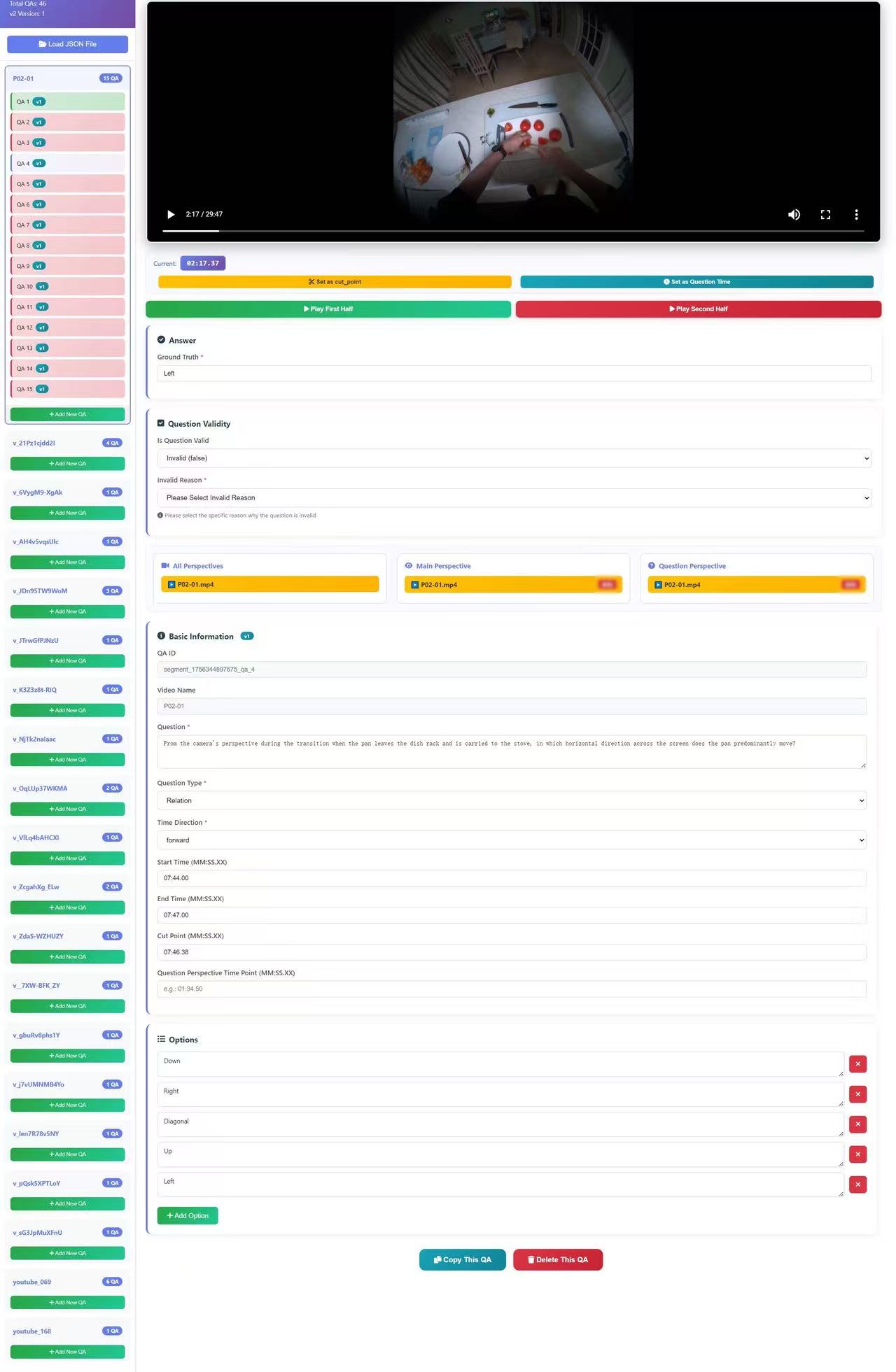}
\vspace{-2mm}
\caption{Repairing tool demonstration. Annotators use it to repair and optimize any attribute of the item as needed.}
\label{fig:repair_tool}
\end{figure*}

\paragraph{Validation and Optimization.}
Next, annotators use the validation tool (cf. Fig.~\ref{fig:validate_tool}) to simulate the actual SCP setting: each question must be answered strictly from the visible portion of the clip.
This process verifies whether the question is answerable, viewpoint-consistent, and aligned with the intended spatial causal reasoning.
Items that fail validation are flagged along with a specific reason, e.g., ``unclear perspective'', ``answerable from visible region'', or ``ambiguous spatial reference''.
If a problematic item is deemed correctable, annotators use the repairing tool (Fig.~\ref{fig:repair_tool}) to edit its question wording, perspective specification, temporal boundary, or answer set.
Through repeated rounds of validation and targeted refinement, annotators progressively resolve inconsistencies and remove invalid items, ultimately yielding the finalized SCP-Bench.

\subsection{Quality Control}
To ensure the overall quality and reliability of SCP-Bench, we employ both a rigorous automated pipeline and a carefully managed human annotation process.
In the \textbf{automated stage}, we leverage GPT-5, one of the strongest commercially available models, and implement a multi-stage filtering workflow that enforces strict constraints on segment selection, clip extraction, and QA generation, thereby ensuring the initial candidate pool is of high quality.
In the \textbf{human annotation stage}, all annotation tasks are conducted by seven well-trained annotators who have undergone extensive instruction on the SCP task definition, selection rules, and evaluation criteria.
As detailed in Sec.~\ref{app:QA_selection}, we design a suite of specialized annotation tools that support efficient inspection, consistent decision-making, and systematic refinement of QA items.
During the \textbf{validation stage}, annotators must provide brief justifications for their decisions, documenting why an item is accepted, rejected, or requires revision.
For items that remain ambiguous, additional annotators re-examine the case and engage in discussion to resolve disagreements.
If consensus cannot be reached, a voting procedure is used to determine whether the item should be retained.
To maintain consistent quality over time, we also conduct periodic acceptance checks and random audits of annotator decisions, ensuring that annotation standards are uniformly upheld throughout the process.

\subsection{Statistics}
Through this carefully designed construction pipeline, we obtain SCP-Bench with 2,500 QA pairs across 1,181 video clips. The benchmark spans four major attributes: question type, causal direction, perspective setting, and scene category, as shown in Table~\ref{tab:statistics_data}.

\begin{table}[t]
\centering
\fontsize{9}{12}\selectfont
\setlength{\tabcolsep}{2.0mm}
\begin{tabular}{lcc}
\toprule
\textbf{Category} & \textbf{Video Clip} & \textbf{QA Instance} \\
\midrule
\rowcolor{lightgrey}\multicolumn{3}{l}{$\bullet$ \textit{Scene Category}} \\
Sports Related & 467 & 964
 \\
Life Records & 289 &  691
 \\
Driving-view Related & 122  & 308
 \\
Animal Related & 85 & 143
 \\
Artistic Performances & 64 & 95
 \\
Factory/Machine Related & 48 & 149
 \\
Others & 106 & 150
 \\
\rowcolor{lightgrey} \multicolumn{3}{l}{$\bullet$ \textit{Perspective}}\\
Ego & 205 & 548 \\
Exo & 943 & 1546 \\
Exo-Exo & 14 & 80 \\
Ego-Exo & 14 & 44 \\
Exo-Ego & 12 & 66 \\
\rowcolor{lightgrey} \multicolumn{3}{l}{$\bullet$ \textit{Causal Direction}}\\
Forward & 1053 & 2037 \\
Backward & 368 & 463 \\
\rowcolor{lightgrey} \multicolumn{3}{l}{$\bullet$ \textit{Question Type}}\\
Planning & 417 & 491 \\
Relation & 641 & 874 \\
Relative Distance & 322 & 376 \\
Spatial State & 270 & 278 \\
Appearance Order & 162 & 167 \\
Relative Speed & 153 & 155  \\
Counting & 117 & 117 \\
Relative Size & 42 & 42 \\
\bottomrule
\end{tabular}
\vspace{-2mm}
\caption{Distribution of benchmarks over the scene category, perspective, Causal direction, and question type. These attributes are combined to form the final dataset, which consists of 1,181 video clips and 2,500 QA pairs.}
\vspace{-3mm}
\label{tab:statistics_data}
\end{table}

\subsection{Human Performance}
Spatial causal reasoning is an intuitive capability that humans routinely exercise in daily life. 
To clearly quantify the gap between current models and human cognition, we conduct a human evaluation on SCP-Bench. 
Participants are presented with the same task setting as the models: only the visible portion of each clip and the corresponding question are provided, and they must select the most appropriate answer from the multiple-choice options. 
The aggregated accuracies provide an estimate of human performance on SCP-Bench.
To ensure the validity and reliability of the human results, we recruit seven participants and enforce a strict ``first-exposure'' rule, i.e., each participant only answers questions they had never seen before, preventing any prior knowledge from influencing the outcomes.

\section{Task Definition}
We define 8 task categories to cover the key variations in spatial causal structure:
\begin{compactitem}
    \item \textbf{Appearance Order}: infers which object or action becomes relevant earlier in the unfolding sequence. 
    \item \textbf{Counting}: tracks cardinality changes driven by underlying spatial dynamics.
    \item \textbf{Planning}: probes short-horizon action tendencies and anticipated movement choices.
    \item \textbf{Relation}: concerns the temporal variation of spatial configurations between objects.
    \item \textbf{Relative Distance}: focuses on how the spatial distance between objects changes over time.
    \item \textbf{Relative Size}: examines the comparative object scale as perceived through the evolving viewpoint.
    \item \textbf{Relative Speed}: concerns how objects move at different speeds and how those speeds change over time.
    \item \textbf{Spatial State}: evaluates how an object’s posture or interaction state is expected to change.
\end{compactitem}

\section{Experimental Details}
\label{app:exp_detail}

\subsection{Model Setup}
\label{app:model_setup}
Our full evaluation process covers 29 models spanning proprietary, open-source, and spatial-reasoning-specialized systems, enabling a broad and comprehensive assessment of current MLLMs.
For proprietary models, we include the latest frontier releases, GPT-5~\cite{openai2025gpt5}, Gemini-2.5-Pro~\cite{comanici2025gemini}, Gemini-2.5-Flash~\cite{comanici2025gemini}, and Claude-Sonnet-4.5~\cite{anthropic2025claudeSonnet4_5}, which represent the most advanced commercially available multimodal reasoning models.
For open-source models, we evaluate a diverse set of leading vision-language systems with different model sizes, including Qwen3-VL~\cite{yang2025qwen3}, Qwen3-Omni~\cite{xu2025qwen3}, InternVL3.5~\cite{wang2025internvl3}, MiniCPM-V-4.5~\cite{yu2025minicpm}, DeepSeek-VL2~\cite{wu2024deepseek}, NVILA~\cite{liu2025nvila}, LLaVA-OneVision~\cite{li2024llava, an2025llava}, and LLaVA-NeXT-Video~\cite{zhang2024video}.
We additionally include Spatial-MLLM~\cite{wu2025spatialmllm} and SpaceR~\cite{ouyang2025spacer}, which are explicitly designed to enhance spatial and relational reasoning.
All models are evaluated using their default inference configurations to ensure consistent and fair comparison.
For models that cannot directly process video inputs, we adopt a standardized preprocessing strategy: sampling one frame per second and feeding the frames sequentially according to their temporal order.
In the following sections, we provide a detailed description of the in-depth analysis settings and present the corresponding experimental results.

\subsection{How Well Do Current Models Perform?}

\subsubsection{Baseline Experiment.}
\label{par:baseline}
To test the current performance of MLLM on SCP-Bench, we strictly follow the setting in our SCP task design:
\begin{compactitem}
    \item \textbf{Baseline}: For single-view questions, the model is given the visible part of a video clip, a question, and options. For multi-view questions, an image indicating the perspective to answer is additionally provided.
\end{compactitem}
The exact prompt templates used in our experiments are shown below for clarity:
\begin{tcolorbox}[colback=black!5!white,colframe=black!75!black,title=Baseline Prompt (Single-View),breakable]
Based on the provided video clip, please answer the following question.
Choose the most appropriate option from the given choices.
(Only answer the choice text without any explanation) 

Question: \{question\} 

Options: \{options\}
\end{tcolorbox}

\begin{tcolorbox}[colback=black!5!white,colframe=black!75!black,title=Baseline Prompt (Multi-View),breakable]
You are given a video and an image from the question perspective.
Based on the events in the video clips, answer the question as if you are the camera in the REFERENCE IMAGE (not the video camera).
Choose the most appropriate option from the given choices.
(Only answer the choice text without any explanation) 

Question: \{question\} 

 Options: \{options\}
\end{tcolorbox}

\begin{table*}[!ht]
\centering
\caption{
Evaluation results for the ``Perception vs. Reasoning Decomposition'' experiment. 
``Avg.'' indicates the overall average accuracy.
\textcolor{red}{Red} and \textcolor{mygreen}{green} indicate performance gains and drops relative to the Baseline setting, respectively.
}
\label{sup_tab:perception_reasoning}
\vspace{-2mm}
\rowcolors{2}{gray!8}{white}
\resizebox{\textwidth}{!}{
\begin{tabular}{lcccccccccc}
\toprule
\textbf{Model} & \cellcolor{lightblue}\bf\textbf{Avg.} & \textbf{Appearance Order} & \textbf{Counting} & \textbf{Planning} & \textbf{Relation} &   \textbf{Relative Distance} & \textbf{Relative Size} & \textbf{Relative Speed} & \textbf{Spatial State} \\
\midrule
\rowcolor{lightgrey} \multicolumn{11}{c}{$\bullet$ \textit{Gold Video}} \\
Qwen3-VL-8B & \cellcolor{lightblue}54.96$_{\textcolor{red}{\uparrow 7.44}}$ & 62.28 & 58.12 & 59.47 & 48.17 & 54.26 & 83.33 & 57.42 & 57.91 \\
InternVL3.5-8B & \cellcolor{lightblue}58.64$_{\textcolor{red}{\uparrow 8.12}}$ & 68.86 & 57.26 & 64.97 & 51.37 & 60.37 & 80.95 & 60.00 & 58.27 \\
MiniCPM-V-4.5 & \cellcolor{lightblue}43.72$_{\textcolor{mygreen}{\downarrow 0.08}}$ & 48.50 & 47.01 & 44.40 & 35.47 & 50.80 & 78.57 & 54.19 & 43.53 \\
\rowcolor{lightgrey} \multicolumn{11}{c}{$\bullet$ \textit{Dense Caption Only}} \\
Qwen3-VL-8B & \cellcolor{lightblue}46.76$_{\textcolor{mygreen}{\downarrow 0.76}}$ & 51.50 & 37.61 & 53.97 & 40.16 & 48.94 & 80.95 & 60.00 & 40.29 \\
InternVL3.5-8B & \cellcolor{lightblue}41.24$_{\textcolor{mygreen}{\downarrow 9.28}}$ & 47.31 & 34.19 & 46.23 & 35.01 & 42.02 & 78.57 & 51.61 & 38.85\\
MiniCPM-V-4.5 & \cellcolor{lightblue}45.76$_{\textcolor{red}{\uparrow 1.96}}$ & 58.08 & 41.88 & 50.51 & 36.04 & 50.00 & 83.33 & 58.06 & 43.88 \\
\rowcolor{lightgrey} \multicolumn{11}{c}{$\bullet$ \textit{Dense Caption Auxiliary}} \\
Qwen3-VL-8B & \cellcolor{lightblue}47.72$_{\textcolor{red}{\uparrow 0.20}}$ & 54.49 & 48.72 & 52.75 & 41.76 & 46.01 & 88.10 & 53.55 & 46.04 \\
InternVL3.5-8B & \cellcolor{lightblue}50.56$_{\textcolor{red}{\uparrow 0.04}}$ & 60.48 & 53.85 & 54.18 & 43.02 & 56.12 & 64.29 & 60.65 & 45.32 \\
MiniCPM-V-4.5 & \cellcolor{lightblue}44.28$_{\textcolor{red}{\uparrow 0.48}}$ & 56.29 & 45.30 & 51.73 & 33.64 & 46.28 & 85.71 & 54.84 & 42.09 \\
\bottomrule
\end{tabular}
}
\end{table*}

\begin{table*}[!ht]
\centering
\caption{
Evaluation results for the ``Single-Frame vs. Multi-Frame Reasoning'' experiment. 
}
\label{tab:singleframe}
\vspace{-2mm}
\rowcolors{2}{gray!8}{white}
\resizebox{\textwidth}{!}{
\begin{tabular}{lcccccccccc}
\toprule
\textbf{Model} & \cellcolor{lightblue}\bf\textbf{Avg.} & \textbf{Appearance Order} & \textbf{Counting} & \textbf{Planning} & \textbf{Relation} &   \textbf{Relative Distance} & \textbf{Relative Size} & \textbf{Relative Speed} & \textbf{Spatial State} \\
\midrule
\rowcolor{lightgrey} \multicolumn{11}{c}{$\bullet$ \textit{Cutpoint Image}} \\
Qwen3-VL-8B & \cellcolor{lightblue}49.52$_{\textcolor{red}{\uparrow 2.00}}$ & 49.70 & 52.14 & 50.92 & 44.62 & 50.27 & 83.33 & 55.48 & 51.80 \\
InternVL3.5-8B & \cellcolor{lightblue}51.92$_{\textcolor{red}{\uparrow 1.40}}$ & 67.07 & 52.99 & 52.14 & 44.16 & 57.18 & 76.19 & 61.29 & 50.36 \\
MiniCPM-V-4.5 & \cellcolor{lightblue}43.80$_{\textcolor{gray}{0.00}}$ & 52.69 & 55.56 & 44.60 & 35.35 & 50.80 & 76.19 & 50.97 & 40.29 \\
Spatial-MLLM & \cellcolor{lightblue}42.04$_{\textcolor{red}{\uparrow 2.28}}$ & 53.46 & 32.73 & 38.90 & 37.30 & 53.55 & 71.43 & 55.69 & 33.33 \\
\bottomrule
\end{tabular}
}
\end{table*}

\begin{table*}[!ht]
\centering
\caption{
Evaluation results for the ``Vision Causal Perception'' experiment. 
}
\label{tab:reversevideo}
\vspace{-2mm}
\rowcolors{2}{gray!8}{white}
\resizebox{\textwidth}{!}{
\begin{tabular}{lcccccccccc}
\toprule
\textbf{Model} & \cellcolor{lightblue}\bf\textbf{Avg.} & \textbf{Appearance Order} & \textbf{Counting} & \textbf{Planning} & \textbf{Relation} & \textbf{Relative Distance} & \textbf{Relative Size} & \textbf{Relative Speed} & \textbf{Spatial State} \\
\midrule
\rowcolor{lightgrey} \multicolumn{11}{c}{$\bullet$ \textit{Reverse Video}} \\
Qwen3-VL-8B & \cellcolor{lightblue}46.76$_{\textcolor{mygreen}{\downarrow 0.76}}$ & 52.69 & 52.14 & 46.84 & 43.94 & 46.54 & 90.48 & 48.39 & 42.45 \\
InternVL3.5-8B & \cellcolor{lightblue}48.56$_{\textcolor{mygreen}{\downarrow 1.96}}$ & 56.89 & 56.41 & 49.49 & 42.33 & 53.19 & 64.29 & 59.35 & 43.53 \\
MiniCPM-V-4.5 & \cellcolor{lightblue}42.24$_{\textcolor{mygreen}{\downarrow 1.56}}$ & 49.10 & 45.30 & 41.96 & 35.13 & 48.67 & 73.81 & 50.97 & 41.37 \\
LLaVA-OneVision-1.5-8B & \cellcolor{lightblue}45.32$_{\textcolor{mygreen}{\downarrow 0.20}}$ & 52.69 & 47.01 & 45.42 & 39.47 & 50.00 & 80.95 & 49.03 & 44.60 \\
\bottomrule
\end{tabular}
}
\end{table*}

\subsection{What Affects Spatial Causal Prediction?}
\subsubsection{Perception vs. Reasoning Decomposition.}
\label{par:perception_reasoning}
This experiment examines the relative importance of visual perception and spatial causal reasoning, aiming to identify the key bottleneck in SCP performance. 
We evaluate Qwen3-VL-8B, InternVL3.5-8B, and MiniCPM-V-4.5 under the following evaluation settings:
\begin{compactitem}
    \item \textbf{Gold Video (Pure Perception)}: The model directly observes the unseen part of the video clip (the gold video), removing the need for causal reasoning.
    \item \textbf{Dense Caption Only (Pure Reasoning)}: Replace video input with its dense caption generated by Tarsier~\cite{wang2024tarsier}, removing the need for video perception.
    \item \textbf{Dense Caption Auxiliary}: The model receives both the video and its dense caption, intending to reduce the perceptual burden.
\end{compactitem}
As illustrated in Table~\ref{sup_tab:perception_reasoning}, the results show that causal reasoning plays a more critical role in SCP performance than raw video perception. 
In addition, supplying dense captions alongside videos leads to further accuracy improvements, suggesting that current models still face limitations in dynamic spatial perception in videos.

\begin{tcolorbox}[colback=black!5!white,colframe=black!75!black,title=Dense Caption Auxiliary Prompt (Single-View),breakable]
Based on the provided video clip, please answer the following question.
Choose the most appropriate option from the given choices.
A dense caption is provided to help understand the video.
(Only answer the choice text without any explanation) \\
Question: \{question\} \\
Options: \{options\}\\
Dense Caption: \{dense caption\}
\end{tcolorbox}

\begin{tcolorbox}[colback=black!5!white,colframe=black!75!black,title=Dense Caption Auxiliary Prompt (Multi-View),breakable]
You are given a video and an image from the question perspective. Based on the events in the VIDEO CLIP, answer the question as if you are the camera in the REFERENCE IMAGE (not the video camera).
Choose the most appropriate option from the given choices.
A dense caption is provided to aid in understanding the video.
(Only answer the choice text without any explanation) \\
Question: \{question\} \\
Options: \{options\}\\
Dense Caption: \{dense caption\}
\end{tcolorbox}

\subsubsection{Single-Frame vs. Multi-Frame Reasoning.}
\label{para:single_multi}
Video contains richer spatial–temporal cues than an image.
To test whether models truly exploit temporal continuity rather than treating frames independently, we compare a single-frame setting with the baseline setting, which uses multiple frames, on Qwen3-VL-8B, InternVL3.5-8B, MiniCPM-V-4.5, and Spatial-MLLM: 
\begin{compactitem}
    \item \textbf{Baseline (Multi-Frame)}: As described in Sec.~\ref{par:baseline}.
    \item \textbf{Cutpoint Image (Single-Frame)}: Replace video input with the frame at the cutpoint.
\end{compactitem}
As shown in Table~\ref{tab:singleframe}, the single-frame input yields slightly better performance across all four models.
This suggests that multi-frame inputs introduce noise rather than useful temporal cues, indicating that models do not effectively use temporal information. 
The single-frame gains further reveal a bias toward image-based perception and limited exploitation of dynamic spatial–temporal signals.

\begin{tcolorbox}[colback=black!5!white,colframe=black!75!black,title=Single Frame Prompt (Single-View),breakable]
Based on the provided image, please answer the following question.
Choose the most appropriate option from the given choices.
(Only answer the choice text without any explanation) 

Question: \{question\} 

Options: \{options\}

\end{tcolorbox}

\begin{tcolorbox}[colback=black!5!white,colframe=black!75!black,title=Single Frame Prompt (Multi-View),breakable]
You are given a reference image (the first) and an image from the question perspective (the second image). Based on the events in the first image, answer the question as if you were the camera in the second image.
Choose the most appropriate option from the given choices.
(Only answer the choice text without any explanation) 

Question: \{question\} 

Options: \{options\}

\end{tcolorbox}

\subsubsection{Vision Causal Perception.}
To examine whether models truly leverage temporal–spatial cues for spatial causal inference, we conduct an experiment in which the input videos are played in reverse.
Specifically, for Qwen3-VL-8B, InternVL3.5-8B, MiniCPM-V-4.5, and LLaVA-OneVision-1.5-8B, we introduce the following variant:
\begin{compactitem}
    \item \textbf{Reverse Video}: Reverse the input video while keeping all other settings identical to the baseline experiment.
\end{compactitem}
As shown in Table~\ref{tab:reversevideo}, model performance exhibits only minor degradation under reversal, suggesting that these systems make limited use of the underlying spatial–causal structure present in video dynamics.

\subsubsection{Visible Range Comparison.}

Unlike standard VideoQA settings, SCP limits models to the visible portion of a clip, making it essential to examine how different visible ranges influence model performance.
We therefore evaluate three input conditions on Qwen3-VL-8B, InternVL3.5-8B, MiniCPM-V-4.5, and Spatial-MLLM:
\begin{compactitem}
    \item \textbf{Baseline (Seen Part)}: As described in Sec.~\ref{par:baseline}.
    \item \textbf{Gold Video (Unseen Part)}: As described in Sec.~\ref{par:perception_reasoning}, the model directly sees the unseen part.
    \item \textbf{Full Video}: The model is given the entire video clip.
\end{compactitem}
As shown in Table~\ref{tab:inputrange}, all models except MiniCPM-V-4.5 achieve their highest accuracy under the gold video setting, followed by full video, with the baseline performing the worst. MiniCPM-V-4.5, however, performs best with full video.
These results indicate that when the answer is directly observable, the task places minimal demands on spatial causal reasoning, leading to higher accuracy.
For most models, the full video introduces additional, potentially distracting content beyond what is strictly necessary, resulting in reduced performance compared with the gold video.

\begin{table*}[!t]
\centering
\caption{
Evaluation results for the ``Visible Range Comparison'' experiment. 
}
\label{tab:inputrange}
\vspace{-2mm}
\rowcolors{2}{gray!8}{white}
\resizebox{\textwidth}{!}{
\begin{tabular}{lcccccccccc}
\toprule
\textbf{Model} & \cellcolor{lightblue}\bf\textbf{Avg.} & \textbf{Appearance Order} & \textbf{Counting} & \textbf{Planning} & \textbf{Relation} & \textbf{Relative Distance} & \textbf{Relative Size} & \textbf{Relative Speed} & \textbf{Spatial State} \\
\midrule
\rowcolor{lightgrey} \multicolumn{11}{c}{$\bullet$ \textit{Baseline}} \\
Qwen3-VL-8B & \cellcolor{lightblue}47.52 & 54.49 & 51.28 & 49.29 & 42.33 & 49.47 & 90.48 & 46.45 & 46.40 \\
InternVL3.5-8B & \cellcolor{lightblue}50.52 & 59.88 & 54.70 & 54.79 & 43.82 & 54.52 & 61.90 & 58.71 & 44.96 \\
MiniCPM-V-4.5 & \cellcolor{lightblue}43.80 & 53.29 & 49.57 & 43.99 & 36.04 & 49.20 & 76.19 & 52.26 & 42.81 \\
Spatial-MLLM & \cellcolor{lightblue}39.76 & 45.51 & 28.21 & 33.81 & 38.33 & 49.73 & 66.67 & 50.97 & 32.37 \\
\rowcolor{lightgrey} \multicolumn{11}{c}{$\bullet$ \textit{Gold Video}} \\
Qwen3-VL-8B & \cellcolor{lightblue}54.96$_{\textcolor{red}{\uparrow 7.44}}$ & 62.28 & 58.12 & 59.47 & 48.17 & 54.26 & 83.33 & 57.42 & 57.91 \\
InternVL3.5-8B & \cellcolor{lightblue}58.64$_{\textcolor{red}{\uparrow 8.12}}$ & 68.86 & 57.26 & 64.97 & 51.37 & 60.37 & 80.95 & 60.00 & 58.27 \\
MiniCPM-V-4.5 & \cellcolor{lightblue}43.72$_{\textcolor{mygreen}{\downarrow 0.08}}$ & 48.50 & 47.01 & 44.40 & 35.47 & 50.80 & 78.57 & 54.19 & 43.53 \\
Spatial-MLLM & \cellcolor{lightblue}41.12$_{\textcolor{red}{\uparrow 1.36}}$ & 37.73 & 24.79 & 34.21 & 32.38 & 47.07 & 66.67 & 49.03 & 28.78 \\
\rowcolor{lightgrey} \multicolumn{11}{c}{$\bullet$ \textit{Full Video}} \\
Qwen3-VL-8B & \cellcolor{lightblue}54.56$_{\textcolor{red}{\uparrow 7.04}}$ & 61.08 & 59.83 & 61.10 & 48.40 & 53.19 & 85.71 & 54.84 & 53.24 \\
InternVL3.5-8B & \cellcolor{lightblue}57.00$_{\textcolor{red}{\uparrow 6.48}}$ & 67.66 & 60.68 & 66.80 & 47.83 & 57.18 & 76.19 & 61.29 & 55.04 \\
MiniCPM-V-4.5 & \cellcolor{lightblue}52.68$_{\textcolor{red}{\uparrow 8.88}}$ & 63.47 & 72.65 & 53.97 & 45.31 & 53.72 & 73.81 & 60.65 & 49.64 \\
Spatial-MLLM & \cellcolor{lightblue}41.12$_{\textcolor{red}{\uparrow 1.36}}$ & 47.10 & 32.73 & 39.92 & 37.76 & 49.20 & 71.43 & 52.69 & 32.73 \\
\bottomrule
\end{tabular}
}
\vspace{-2mm}
\end{table*}

\begin{table*}[!t]
\centering
\caption{
Evaluation results for the ``Is Visual Information Necessary?'' experiment. 
}
\label{tab:visual_information}
\vspace{-2mm}
\rowcolors{2}{gray!8}{white}
\resizebox{\textwidth}{!}{
\begin{tabular}{lcccccccccc}
\toprule
\textbf{Model} & \cellcolor{lightblue}\bf\textbf{Avg.} & \textbf{Appearance Order} & \textbf{Counting} & \textbf{Planning} & \textbf{Relation} &   \textbf{Relative Distance} & \textbf{Relative Size} & \textbf{Relative Speed} & \textbf{Spatial State} \\
\midrule
\rowcolor{lightgrey} \multicolumn{11}{c}{$\bullet$ \textit{Pure Text}} \\
Qwen3-VL-8B & \cellcolor{lightblue}42.00$_{\textcolor{mygreen}{\downarrow 5.52}}$ & 49.10 & 35.04 & 47.25 & 34.67 & 46.28 & 69.05 & 40.65 & 45.32 \\
InternVL3.5-8B & \cellcolor{lightblue}44.72$_{\textcolor{mygreen}{\downarrow 5.80}}$ & 50.90 & 42.74 & 48.88 & 36.61 & 52.66 & 73.81 & 51.61 & 41.01 \\
MiniCPM-V-4.5 & \cellcolor{lightblue}43.32$_{\textcolor{mygreen}{\downarrow 0.48}}$ & 53.29 & 44.44 & 46.03 & 35.01 & 51.60 & 69.05 & 52.90 & 37.77 \\
LLaVA-OneVision-1.5-8B & \cellcolor{lightblue}40.20$_{\textcolor{mygreen}{\downarrow 5.32}}$ & 43.11 & 41.03 & 42.16 & 33.41 & 47.87 & 73.81 & 38.71 & 41.37 \\
\rowcolor{lightgrey} \multicolumn{11}{c}{$\bullet$ \textit{Dense Caption Only}} \\
Qwen3-VL-8B & \cellcolor{lightblue}46.76$_{\textcolor{mygreen}{\downarrow 0.76}}$ & 51.50 & 37.61 & 53.97 & 40.16 & 48.94 & 80.95 & 60.00 & 40.29 \\
InternVL3.5-8B & \cellcolor{lightblue}41.24$_{\textcolor{mygreen}{\downarrow 9.28}}$ & 47.31 & 34.19 & 46.23 & 35.01 & 42.02 & 78.57 & 51.61 & 38.85 \\
MiniCPM-V-4.5 & \cellcolor{lightblue}45.76$_{\textcolor{red}{\uparrow 1.96}}$ & 58.08 & 41.88 & 50.51 & 36.04 & 50.00 & 83.33 & 58.06 & 43.88 \\
LLaVA-OneVision-1.5-8B & \cellcolor{lightblue}43.12$_{\textcolor{mygreen}{\downarrow 2.40}}$ & 49.70 & 29.91 & 50.31 & 36.04 & 44.41 & 69.05 & 53.55 & 42.81 \\

\bottomrule
\end{tabular}
}
\end{table*}

\subsubsection{Is Visual Information Necessary?}
Modern MLLMs exhibit strong text-based abilities, motivating an examination of whether visual information is needed for SCP.
To investigate this, we consider two conditions: answering SCP questions using their internal priors and only a textual description of the video.
We evaluate Qwen3-VL-8B, InternVL3.5-8B, MiniCPM-V-4.5, and LLaVA-OneVision-1.5-8B under the following settings:
\begin{compactitem}
    \item \textbf{Pure Text (Internal Priors)}: Answering only the textual question without any visual information.
    \item \textbf{Dense Caption Only (Textual Information)}: As described in Sec.~\ref{par:perception_reasoning}.
    \item \textbf{Baseline (Visual Information)}: The baseline setting as described in Sec.~\ref{par:baseline}.
\end{compactitem}
The results in Table~\ref{tab:visual_information} indicate that when models rely solely on their intrinsic prior knowledge (i.e., pure text input), their performance drops substantially.
Although supplying dense captions leads to noticeable improvements, the gains fall short of the baseline that uses video inputs.
These findings suggest that textual priors alone are insufficient for SCP, and even dense captions cannot fully capture the richness, continuity, and fine-grained cues encoded in visual observations.

\begin{tcolorbox}[colback=black!5!white,colframe=black!75!black,title=PureText Prompt,breakable]
Please answer the following question.
Choose the most appropriate option from the given choices.
(Only answer the choice text without any explanation) 

Question: \{question\} 

Options: \{options\}
\end{tcolorbox}

\begin{tcolorbox}[colback=black!5!white,colframe=black!75!black,title=Dense Caption Only Prompt (Single-View),breakable]
Based on the provided dense caption description, please answer the following question.
Choose the most appropriate option from the given choices.
(Only answer the choice text without any explanation) 

Question: \{question\} 

Options: \{options\}

Dense Caption: \{dense caption\}
\end{tcolorbox}

\begin{tcolorbox}[colback=black!5!white,colframe=black!75!black,title=Dense Caption Only Prompt (Multi-View),breakable]
You are given a video dense caption and a reference image dense caption describing another viewpoint.
Based on the dense captions, answer
the question as if you were the camera in the reference image.
Choose the most appropriate option from the given choices.
(Only answer the choice text without any explanation) 

Question: \{question\} 

Options: \{options\}

Video Dense Caption: \{video dense caption\}

Image Dense Caption: \{image dense caption\}

\end{tcolorbox}

\begin{table*}[!ht]
\centering
\caption{
Evaluation results for the ``Model Scale-up Effect Analysis'' experiment. 
}
\label{tab:model_scale}
\vspace{-2mm}
\rowcolors{2}{gray!8}{white}
\resizebox{\textwidth}{!}{
\begin{tabular}{lcccccccccc}
\toprule
\textbf{Model} & \cellcolor{lightblue}\bf\textbf{Avg.} & \textbf{Appearance Order} & \textbf{Counting} & \textbf{Planning} & \textbf{Relation} &   \textbf{Relative Distance} & \textbf{Relative Size} & \textbf{Relative Speed} & \textbf{Spatial State} \\
\midrule
\rowcolor{lightgrey} \multicolumn{11}{c}{$\bullet$ \textit{Qwen3-VL Series}} \\
Qwen3-VL-2B & \cellcolor{lightblue}43.04 & 41.92 & 42.74 & 45.01 & 40.85 & 44.41 & 59.52 & 47.10 & 40.65 \\
Qwen3-VL-4B & \cellcolor{lightblue}50.24 & 58.68 & 47.86 & 52.55 & 44.51 & 55.05 & 90.48 & 60.00 & 42.09 \\
Qwen3-VL-8B & \cellcolor{lightblue}47.52 & 54.49 & 51.28 & 49.29 & 42.33 & 49.47 & 90.48 & 46.45 & 46.40 \\
Qwen3-VL-30B-A3B & \cellcolor{lightblue}54.16 & 65.27 & 52.14 & 54.79 & 46.22 & 56.65 & 85.71 & 66.45 & 57.19 \\
Qwen3-VL-32B & \cellcolor{lightblue}56.84 & 59.88 & 51.28 & 58.66 & 52.63 & 57.98 & 90.48 & 67.10 & 55.04 \\
Qwen3-VL-235B-A22B & \cellcolor{lightblue}61.04 & 67.07 & 54.70 & 60.90 & 55.03 & 63.03 & 97.62 & 74.84 & 63.31 \\

\rowcolor{lightgrey} \multicolumn{11}{c}{$\bullet$ \textit{InternVL3.5 Series}} \\
InternVL3.5-1B & \cellcolor{lightblue}35.16 & 29.94 & 43.59 & 37.07 & 30.66 & 39.36 & 38.10 & 49.03 & 31.65 \\
InternVL3.5-2B & \cellcolor{lightblue}40.84 & 44.91 & 44.44 & 45.01 & 37.07 & 40.16 & 61.90 & 46.45 & 35.97 \\
InternVL3.5-4B & \cellcolor{lightblue}46.12 & 57.49 & 47.86 & 44.60 & 42.79 & 47.61 & 76.19 & 49.68 & 43.17 \\
InternVL3.5-8B & \cellcolor{lightblue}50.52 & 59.88 & 54.70 & 54.79 & 43.82 & 54.52 & 61.90 & 58.71 & 44.96 \\
InternVL3.5-14B & \cellcolor{lightblue}50.28 & 59.88 & 57.26 & 54.79 & 42.56 & 54.26 & 69.05 & 63.23 & 42.45 \\
InternVL3.5-30B-A3B & \cellcolor{lightblue}52.48 & 65.87 & 52.99 & 56.01 & 43.25 & 56.12 & 80.95 & 67.10 & 49.64 \\
InternVL3.5-38B & \cellcolor{lightblue}53.56 & 62.28 & 53.85 & 56.01 & 46.34 & 57.98 & 90.48 & 65.81 & 48.20 \\
InternVL3.5-241B-A28B & \cellcolor{lightblue}56.96 & 67.07 & 60.68 & 61.10 & 46.11 & 60.37 & 90.48 & 68.39 & 60.07 \\

\bottomrule
\end{tabular}
}
\end{table*}

\begin{table*}[!ht]
\centering
\caption{
Evaluation results for the ``CoT Reasoning'' experiment. 
}
\label{tab:cot}
\vspace{-2mm}
\rowcolors{2}{gray!8}{white}
\resizebox{\textwidth}{!}{
\begin{tabular}{lcccccccccc}
\toprule
\textbf{Model} & \cellcolor{lightblue}\bf\textbf{Avg.} & \textbf{Appearance Order} & \textbf{Counting} & \textbf{Planning} & \textbf{Relation} &   \textbf{Relative Distance} & \textbf{Relative Size} & \textbf{Relative Speed} & \textbf{Spatial State} \\
\midrule
\rowcolor{lightgrey} \multicolumn{11}{c}{$\bullet$ \textit{Vanilla CoT}} \\
Qwen3-VL-8B & \cellcolor{lightblue}51.52$_{\textcolor{red}{\uparrow 4.00}}$ & 63.47 & 50.43 & 51.53 & 44.39 & 54.52 & 90.48 & 61.29 & 51.80 \\
InternVL3.5-8B & \cellcolor{lightblue}50.36$_{\textcolor{mygreen}{\downarrow 0.16}}$ & 59.28 & 51.28 & 49.29 & 43.02 & 56.65 & 61.90 & 65.16 & 51.08 \\
MiniCPM-V-4.5 & \cellcolor{lightblue}44.88$_{\textcolor{red}{\uparrow 1.08}}$ & 55.09 & 40.17 & 48.68 & 35.47 & 53.19 & 73.81 & 59.35 & 39.93 \\
LLaVA-OneVision-1.5-8B & \cellcolor{lightblue}42.88$_{\textcolor{mygreen}{\downarrow 2.64}}$ & 48.50 & 39.32 & 46.44 & 33.30 & 53.99 & 69.05 & 54.19 & 39.57 \\
\bottomrule
\end{tabular}
}
\end{table*}

\subsection{How to Improve Spatial Causal Prediction?}

\subsubsection{Model Scale-up Effect Analysis.}
Increasing model size is a direct way to improve performance. 
To examine how parameter scale affects SCP, we evaluate two model families with rich size variations, Qwen3-VL and InternVL-3.5, covering 14 models as listed in Table~\ref{tab:model_scale}. 
Overall, scaling up is an effective strategy for SCP. 
The largest Qwen3-VL-235B-A22B model achieves an improvement of 18\% over the smallest Qwen3-VL-2B model. 
Similarly, the largest InternVL3.5-241B-A28B model outperforms the smallest InternVL-1B model by 21.81\%.
However, scaling is not strictly monotonic among the smaller models.
For example, Qwen3-VL-8B performs 2.72\% worse than Qwen3-VL-4B. 
This suggests that the benefits of scaling become stable only when the increase in parameters reaches a more substantial range.

\subsubsection{CoT Reasoning.}
To investigate whether chain of thought (CoT) prompting can improve performance on SCP, we apply a vanilla CoT setting to Qwen3-VL-8B, InternVL3.5-8B, MiniCPM-V-4.5, and LLaVA-OV-1.5-8B:
\begin{compactitem}
    \item \textbf{Vanilla CoT}: Augment the prompt with ``Think step by step'' to encourage the model to generate its chain of thought reasoning.
\end{compactitem}
As shown in Table~\ref{tab:cot}, Qwen3-VL-8B gains about 4\% in accuracy, while the other three models show no clear improvement. This indicates that vanilla CoT prompting does not consistently enhance SCP performance.

\begin{tcolorbox}[colback=black!5!white,colframe=black!75!black,title=Vanilla CoT Prompt (Single-View),breakable]
Based on the provided video clip, please think step by step to answer the following question.
Show your reasoning process explicitly, and at the end, provide your final choice marked as `Final Answer:'.
Your final answer should contain only the option text, not its letter.

Question: \{question\} 

Options: \{options\}

\end{tcolorbox}

\begin{tcolorbox}[colback=black!5!white,colframe=black!75!black,title=Vanilla CoT Prompt (Multi-View),breakable]
You are given a video and a reference image showing another viewpoint.
Based on the events in the VIDEO CLIP, reason step by step as if you were the camera in the REFERENCE IMAGE, not the video camera.
Show your reasoning process explicitly, and at the end, provide your final choice marked as `Final Answer:'. 
Your final answer should contain only the option text, not its letter.

Question: \{question\} 

Options: \{options\}

\end{tcolorbox}

\begin{tcolorbox}[colback=black!5!white,colframe=black!75!black,title=InternVL3.5 Thinking Mode System Prompt,breakable]
You are an AI assistant that rigorously follows this response protocol:

1. First, conduct a detailed analysis of the question. Consider different angles, potential solutions, and reason through the problem step-by-step. Enclose this entire thinking process within \texttt{<think>} and \texttt{</think>} tags.

2. After the thinking section, provide a clear, concise, and direct answer to the user's question. Separate the answer from the think section with a newline.

Ensure that the thinking process is thorough but remains focused on the query. The final answer should be standalone and not reference the thinking section.

\end{tcolorbox}

\begin{table*}[!ht]
\centering
\caption{
Evaluation results for the ``Self-Think Reasoning'' experiment. 
}
\label{tab:thinking_mode}
\rowcolors{2}{gray!8}{white}
\resizebox{\textwidth}{!}{
\begin{tabular}{lcccccccccc}
\toprule
\textbf{Model} & \cellcolor{lightblue}\bf\textbf{Avg.} & \textbf{Appearance Order} & \textbf{Counting} & \textbf{Planning} & \textbf{Relation} &   \textbf{Relative Distance} & \textbf{Relative Size} & \textbf{Relative Speed} & \textbf{Spatial State} \\
\midrule
\rowcolor{lightgrey} \multicolumn{11}{c}{$\bullet$ \textit{Self-Thinking}} \\
Qwen3-VL-8B & \cellcolor{lightblue}47.44$_{\textcolor{mygreen}{\downarrow 0.08}}$ & 55.69 & 45.30 & 50.10 & 38.79 & 51.86 & 83.33 & 58.71 & 48.20 \\
InternVL3.5-8B & \cellcolor{lightblue}47.84$_{\textcolor{mygreen}{\downarrow 2.68}}$ & 59.28 & 51.28 & 51.12 & 39.36 & 53.46 & 66.67 & 54.19 & 46.40 \\
MiniCPM-V-4.5 & \cellcolor{lightblue}42.72$_{\textcolor{mygreen}{\downarrow 1.08}}$ & 49.10 & 48.72 & 40.94 & 35.35 & 50.53 & 69.05 & 50.32 & 43.88 \\
\bottomrule
\end{tabular}
}
\end{table*}

\begin{table*}[!ht]
\centering
\caption{
Evaluation results for the ``Perception Enhancement Strategy'' experiment. 
}
\label{tab:perception_enhancement}
\rowcolors{2}{gray!8}{white}
\resizebox{\textwidth}{!}{
\begin{tabular}{lcccccccccc}
\toprule
\textbf{Model} & \cellcolor{lightblue}\bf\textbf{Avg.} & \textbf{Appearance Order} & \textbf{Counting} & \textbf{Planning} & \textbf{Relation} &   \textbf{Relative Distance} & \textbf{Relative Size} & \textbf{Relative Speed} & \textbf{Spatial State} \\
\midrule
\rowcolor{lightgrey} \multicolumn{11}{c}{$\bullet$ \textit{Dense Caption Auxiliary}} \\
Qwen3-VL-8B & \cellcolor{lightblue}47.72$_{\textcolor{red}{\uparrow 0.20}}$ & 54.49 & 48.72 & 52.75 & 41.76 & 46.01 & 88.10 & 53.55 & 46.04 \\
InternVL3.5-8B & \cellcolor{lightblue}50.56$_{\textcolor{red}{\uparrow 0.04}}$ & 60.48 & 53.85 & 54.18 & 43.02 & 56.12 & 64.29 & 60.65 & 45.32 \\
MiniCPM-V-4.5 & \cellcolor{lightblue}44.28$_{\textcolor{red}{\uparrow 0.48}}$ & 56.29 & 45.30 & 51.73 & 33.64 & 46.28 & 85.71 & 54.84 & 42.09 \\
LLaVA-OneVision-1.5-8B & \cellcolor{lightblue}45.80$_{\textcolor{red}{\uparrow 0.28}}$ & 53.89 & 46.15 & 49.08 & 39.13 & 47.61 & 83.33 & 54.84 & 42.81 \\
\rowcolor{lightgrey} \multicolumn{11}{c}{$\bullet$ \textit{Spatial Interaction Graph}} \\
Qwen3-VL-8B & \cellcolor{lightblue}51.76$_{\textcolor{red}{\uparrow 4.24}}$ & 69.46 & 46.15 & 52.95 & 45.08 & 55.05 & 78.57 & 70.32 & 43.53 \\
InternVL3.5-8B & \cellcolor{lightblue}48.80$_{\textcolor{mygreen}{\downarrow 1.72}}$ & 59.88 & 45.30 & 49.49 & 38.67 & 55.85 & 78.57 & 67.10 & 50.00 \\
MiniCPM-V-4.5 & \cellcolor{lightblue}43.88$_{\textcolor{red}{\uparrow 0.08}}$ & 59.88 & 42.74 & 47.66 & 32.49 & 48.14 & 76.19 & 59.35 & 44.60 \\
LLaVA-OneVision-1.5-8B & \cellcolor{lightblue}42.16$_{\textcolor{mygreen}{\downarrow 3.36}}$ & 52.69 & 44.44 & 43.18 & 33.87 & 47.07 & 76.19 & 49.68 & 43.17 \\
\bottomrule
\end{tabular}
}
\end{table*}

\subsubsection{Self-Think Reasoning.}
Besides chain-of-thought prompting, some models are trained on data with explicit reasoning traces, which encourages latent internal reasoning during inference time.
To evaluate whether self-thinking improves SCP performance, we configure the thinking modes of different models as follows:

\begin{compactitem}
    \item \textbf{Self-Thinking}: each model is run with its official thinking mode. 
    Qwen3-VL-8B uses the Qwen3-VL-8B-Thinking variant; InternVL3.5-8B enables thinking via its system prompt; 
    MiniCPM-V-4.5 activates thinking with \texttt{enable\_thinking=true}.
\end{compactitem}

The results in Table~\ref{tab:thinking_mode} show that enabling self-think leads to a slight performance drop compared with the baseline. 
This suggests that the current self-think mechanisms do not provide useful spatial causal reasoning signals and may even introduce additional noise that harms performance.

\begin{table*}[!ht]
\centering
\caption{
Evaluation results for the ``Causal Prediction Enhancement Strategy'' experiment. 
}
\label{tab:prediction}
\vspace{-2mm}
\rowcolors{2}{gray!8}{white}
\resizebox{\textwidth}{!}{
\begin{tabular}{lcccccccccc}
\toprule
\textbf{Model} & \cellcolor{lightblue}\bf\textbf{Avg.} & \textbf{Appearance Order} & \textbf{Counting} & \textbf{Planning} & \textbf{Relation} &   \textbf{Relative Distance} & \textbf{Relative Size} & \textbf{Relative Speed} & \textbf{Spatial State} \\
\midrule
\rowcolor{lightgrey} \multicolumn{11}{c}{$\bullet$ \textit{Text Scaffold}} \\
Qwen3-VL-8B & \cellcolor{lightblue}61.23$_{\textcolor{red}{\uparrow 13.71}}$ & 64.29 & 56.67 & 67.44 & 54.48 & 64.08 & 91.89 & 64.06 & 56.25 \\
InternVL3.5-8B & \cellcolor{lightblue}61.72$_{\textcolor{red}{\uparrow 11.20}}$ & 64.29 & 57.78 & 67.44 & 53.25 & 65.49 & 91.89 & 67.97 & 59.13 \\
MiniCPM-V-4.5 & \cellcolor{lightblue}60.15$_{\textcolor{red}{\uparrow 16.35}}$ & 58.16 & 52.22 & 66.28 & 51.32 & 64.08 & 97.30 & 67.19 & 59.62 \\
LLaVA-OneVision-1.5-8B & \cellcolor{lightblue}59.77$_{\textcolor{red}{\uparrow 14.52}}$ & 55.10 & 53.33 & 66.05 & 50.09 & 62.68 & 89.19 & 71.09 & 62.02 \\
\rowcolor{lightgrey} \multicolumn{11}{c}{$\bullet$ \textit{Image Scaffold}} \\
Qwen3-VL-8B & \cellcolor{lightblue}49.21$_{\textcolor{red}{\uparrow 1.69}}$ & 46.94 & 36.67 & 57.27 & 43.76 & 48.24 & 91.89 & 50.78 & 46.63 \\
InternVL3.5-8B & \cellcolor{lightblue}51.60$_{\textcolor{red}{\uparrow 1.08}}$ & 52.04 & 43.33 & 58.20 & 47.10 & 52.46 & 72.97 & 56.25 & 45.67 \\
MiniCPM-V-4.5 & \cellcolor{lightblue}43.64$_{\textcolor{mygreen}{\downarrow 0.16}}$ & 48.98 & 53.33 & 44.57 & 36.91 & 46.48 & 81.08 & 50.78 & 38.46 \\
LLaVA-OneVision-1.5-8B & \cellcolor{lightblue}46.13$_{\textcolor{red}{\uparrow 0.61}}$ & 56.12 & 35.56 & 52.42 & 38.84 & 48.24 & 83.78 & 45.31 & 43.75 \\
\rowcolor{lightgrey} \multicolumn{11}{c}{$\bullet$ \textit{Video Scaffold}} \\
Qwen3-VL-8B & \cellcolor{lightblue}52.63$_{\textcolor{red}{\uparrow 5.11}}$ & 51.02 & 45.56 & 57.51 & 48.86 & 51.76 & 94.59 & 52.34 & 50.48 \\
InternVL3.5-8B & \cellcolor{lightblue}53.38$_{\textcolor{red}{\uparrow 2.86}}$ & 56.12 & 47.78 & 59.35 & 47.98 & 53.17 & 86.49 & 57.03 & 49.04 \\
MiniCPM-V-4.5 & \cellcolor{lightblue}44.29$_{\textcolor{red}{\uparrow 0.49}}$ & 47.96 & 43.33 & 45.03 & 37.79 & 50.35 & 83.78 & 51.56 & 39.42 \\
LLaVA-OneVision-1.5-8B & \cellcolor{lightblue}47.86$_{\textcolor{red}{\uparrow 2.34}}$ & 51.02 & 45.56 & 51.96 & 41.65 & 52.11 & 81.08 & 46.88 & 44.71 \\
\bottomrule
\end{tabular}
}
\end{table*}

\subsubsection{Perception Enhancement Strategy.}
Although perception is not the main bottleneck for SCP, previous experiments indicate that models still struggle with some dynamic spatial–temporal cues. 
We therefore explore 2 strategies that modestly enhance perceptual input:
\begin{compactitem}
    \item \textbf{Dense Caption Auxiliary}: As described in Sec.~\ref{par:perception_reasoning}.
    \item \textbf{Spatial Interaction Graph}: The model is guided to identify key objects and the key background elements, from which a spatial relation graph and an interaction graph are constructed. The model answers the question based on this structured spatial representation.
\end{compactitem}
The results in Table~\ref{tab:perception_enhancement} show that both enhancements yield only marginal improvements. This indicates that perception-level augmentation alone is insufficient to meaningfully improve SCP performance.

\begin{tcolorbox}[colback=black!5!white,colframe=black!75!black,title=Spatial-Interaction Graph Guiding Prompt,breakable]

Based on the provided video clip, you should \texttt{\{temporal\_direction\}} to answer the question. \texttt{\{view\_text\}} Please answer the question based on the following framework and think step by step to analyze through each part.
Your final answer should contain only the option text, not its letter.

\textbf{Framework:}

\textbf{Part 1.} Carefully analyze the video to identify the key spatial changing objects and their affected environmental elements that need to be addressed in answering the questions. You can construct triples of the form:
\texttt{\{
  "changing\_objects": ["..."],
  "affected\_environmental\_elements": ["..."]
\}}
Only list the objects and environmental elements that need to be addressed in answering the question.

\textbf{Part 2.} Construct the spatial graph that describes the spatial relations among the entities, which is helpful for reasoning about the question. You can construct triples of the form: 
\texttt{\{"subject": "...", "relation": "...", "object": "..."\}} 
to describe the spatial relations and trends of changes during the video clip.

\textbf{Part 3.} Construct the interaction graph that describes the interaction relations among the entities, which is helpful for reasoning about the question. You can construct triples of the form: 
\texttt{\{"subject": "...", "action": "...", "object": "..."\}} 
to describe the interaction during the video clip.

\textbf{Part 4.} \texttt{\{temporal\_text\}}

\textbf{Part 5.} Based on previous thinking and constructed, think through and choose the most spatio-temporally consistent option.

\textbf{Rules:}
\begin{compactitem}
    \item Show your reasoning process explicitly, and at the end provide your final choice marked as 'Final Answer:'.
    \item Your final answer should contain only the option text, not its letter.
    \item Don't get caught up in repetitive reasoning.
\end{compactitem}

Question: \{question\} 

Options: \{options\}

\end{tcolorbox}

\begin{table*}[!ht]
  \caption{
  Comparison with existing benchmarks. 
  \texttt{Modality} denotes the input type. 
  \texttt{Dynamic/Static} indicates whether the scene content itself undergoes temporal changes, rather than mere camera movement. 
  \texttt{View Type} specifies whether the questions involve single or multiple viewpoints. 
  \texttt{Perspective} denotes whether the benchmark includes explicitly designed ego (first-person) and exo (third-person) perspective settings.
  \texttt{Causal Reasoning} indicates whether the benchmark requires inferring outcomes or states driven by causal dependencies.
  \texttt{Seen/Unseen} reflects whether the queried information is directly observable within the given visual content. 
  }
  \label{tab:benchmark_comparison}
  \vspace{-2mm}
  \centering
  \fontsize{9}{10}\selectfont
  \setlength{\tabcolsep}{1.4mm}
  \begin{tabular}{@{}lccccccccc}
    \toprule
      \multirow{2}{*}{\textbf{Benchmark}}
      & \multirow{2}{*}{\textbf{QA pairs}}
      & \multirow{2}{*}{\textbf{Modality}}
      & \multirow{2}{*}{\textbf{Dynamic/Static}}
      & \multicolumn{2}{c}{\textbf{View Type}}
      & \multicolumn{2}{c}{\textbf{Perspective}}
      & \multirow{2}{*}{\textbf{Causal Reasoning}}
      & \multirow{2}{*}{\textbf{Seen/Unseen}} \\
      \cmidrule(lr){5-6}
        \cmidrule(lr){7-8}
        & & & 
        & \textbf{Single} & \textbf{Multi}
        & \textbf{Ego} & \textbf{Exo} 
        & \\
    \midrule
    3DSRBench~\cite{ma20253dsrbench} & 3,772 & Image & Static & \checkmark  & \crossmark & \crossmark  &  \checkmark & \crossmark & Seen\\
    InternSpatial-Bench~\cite{deng2025internspatial} &  6,008 & Image & Static & \checkmark  & \crossmark & \crossmark  &  \checkmark & \crossmark & Seen \\
    OmniSpatial~\cite{jia2025omnispatial} & $\sim$8,400 & Image & Static & \checkmark & \crossmark & \checkmark & \checkmark  & \crossmark & Seen\\
    Spatial457~\cite{wang2025spatial457} & 23,752 & Image & Static & \checkmark & \crossmark & \multicolumn{2}{c}{\cellcolor{lightgray}\textit{Undeclared}}  & \crossmark & Seen\\
    All-Angles Bench~\cite{yeh2025seeing} & $\sim$2,100 & Image & Static  & \crossmark & \checkmark & \multicolumn{2}{c}{\cellcolor{lightgray}\textit{Undeclared}} & \crossmark  & Seen \\
    EmbSpatial-Bench~\cite{du2024embspatial} & 3,640 & Image & Static & \checkmark  & \crossmark & \checkmark & \crossmark & \crossmark &  Seen \\
    MMSI-Bench~\cite{yang2025mmsi} & 1,000 & Image & Static & \crossmark & \checkmark  & \multicolumn{2}{c}{\cellcolor{lightgray}\textit{Undeclared}} & \crossmark & Seen\\
    MindCube~\cite{yin2025spatial} & 21,154 & Image & Static & \crossmark & \checkmark & \multicolumn{2}{c}{\cellcolor{lightgray}\textit{Undeclared}}  & \crossmark & Unseen \\
    VSI-Bench~\cite{yang2025thinking} & 5,130 & Video & Static & \checkmark & \crossmark & \checkmark & \crossmark & \crossmark & Seen\\
    VLM4D~\cite{zhou2025vlm4d} & $\sim$1,800 & Video & Dynamic & \checkmark & \crossmark &  \checkmark & \checkmark & \crossmark & Seen \\
    STI-Bench~\cite{li2025sti} & 2,064 & Video & Dynamic & \checkmark & \crossmark & \checkmark & \checkmark & \crossmark &  Seen\\
    DSI-Bench~\cite{zhang2025dsi} & $\sim$1,700 & Video & Dynamic & \checkmark & \crossmark & \multicolumn{2}{c}{\cellcolor{lightgray}\textit{Undeclared}} & \crossmark &  Seen\\
    \midrule
    SCP-Bench (Ours) & 2,500  & Video &  Dynamic & \checkmark & \checkmark & \checkmark & \checkmark & \checkmark&  Unseen\\
    \bottomrule
  \end{tabular}
\end{table*}

\subsubsection{Causal Prediction Enhancement Strategy.}
\begin{figure*}[!t]
\centering
\includegraphics[width=0.95\textwidth]{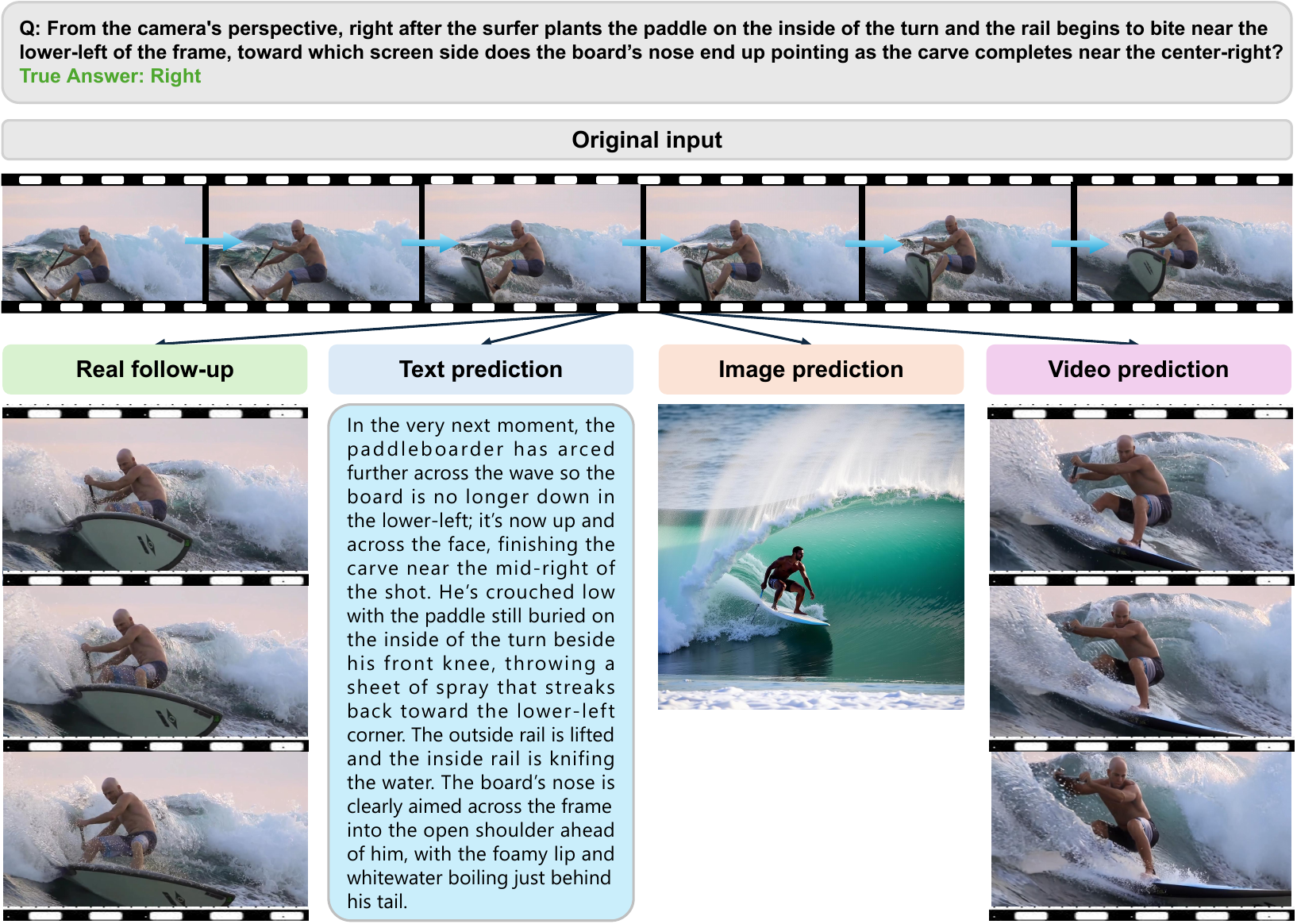}
\vspace{-2mm}
\caption{
Example of causal prediction enhancement. 
Here we present a comparative analysis of a case study, showing the actual subsequent video and the causal prediction results for text, image, and video modalities.
}
\label{fig:causal_prediction}
\end{figure*}

To investigate whether spatial causal prediction scaffolds can mitigate the reasoning bottleneck in SCP, and which modality provides the greatest benefit, 
we evaluate models on a simplified subset of the benchmark that is single-view and requires predicting future states. 
The experiments are performed on Qwen3-VL-8B, InternVL3.5-8B, MiniCPM-V-4.5 and LLaVA-OneVision-1.5-8B in 3 modalities:
\begin{compactitem}
    \item \textbf{Text Scaffold}: Using GPT-5~\cite{openai2025gpt5}, which performs best on SCP, we generate a detailed description of the future that is required to answer the question.
    \item \textbf{Image Scaffold}: Using Flux.1-dev-12B~\cite{flux2024}, we generate a predictive future frame directly based on the textual future description.
    \item \textbf{Video Scaffold}: Using Wan2.2-TI2V-5B~\cite{wan2025wan}, we attempt to generate a video approximation of the unseen portion by starting from the cutpoint frame and guiding the synthesis with the textual future description.
\end{compactitem}
Fig.~\ref{fig:causal_prediction} shows an example using scaffolds.
Each scaffold is provided as auxiliary information to support the model in answering the question.
In Table~\ref{tab:prediction}, all four models show performance gains, with textual scaffolds yielding the largest improvement, followed by video scaffolds. 
Image scaffolds offer the weakest benefit.
Textual scaffolds offer the most complete and explicit description of the future. 
In contrast, image and video scaffolds can only approximate the real scene to a limited extent because they do not capture the full textual detail.
Video scaffolds, however, perform better than image scaffolds because they provide richer temporal and spatial cues.

\begin{tcolorbox}[colback=black!5!white,colframe=black!75!black,title=Text Scaffold Prompt,breakable]
Based on the provided video clips and the future description, please answer the following question.
Choose the most appropriate option from the given choices.
(Only answer the choice text without any explanation) 

Question: \{question\} 

Options: \{options\}

Future Description: \{future\_description\}

\end{tcolorbox}

\begin{tcolorbox}[colback=black!5!white,colframe=black!75!black,title=Image Scaffold Prompt,breakable]
Based on the provided video clips and a future image after the video, please answer the following question.
Choose the most appropriate option from the given choices.
(Only answer the choice text without any explanation) 

Question: \{question\} 

Options: \{options\}

\end{tcolorbox}

\begin{tcolorbox}[colback=black!5!white,colframe=black!75!black,title=Video Scaffold Prompt,breakable]
Based on the provided video clips and the generated future video clips, please answer the following question.
Choose the most appropriate option from the given choices.
(Only answer the choice text without any explanation) 

Question: \{question\} 

Options: \{options\}

\end{tcolorbox}

\begin{table*}[!ht]
\centering
\caption{
Results of the ``Extended Causal Consistency'' experiment, where models are evaluated by how often they misclassify the correct answer as the least plausible option. 
Lower values indicate stronger causal consistency. ``Avg.'' denotes the overall average error rate.
}
\label{tab:extended_causal_consistency}
\vspace{-2mm}
\rowcolors{2}{gray!8}{white}
\resizebox{\textwidth}{!}{
\begin{tabular}{lcccccccccc}
\toprule
\textbf{Model} & \cellcolor{lightblue}\textbf{Avg.}$\downarrow$ & \textbf{Appearance Order} & \textbf{Counting} & \textbf{Planning} & \textbf{Relation} &   \textbf{Relative Distance} & \textbf{Relative Size} & \textbf{Relative Speed} & \textbf{Spatial State} \\
\midrule
\rowcolor{lightgrey} \multicolumn{11}{c}{$\bullet$ \textit{Misclassifying the Correct Answer as Least Plausible}} \\
Qwen3-VL-8B & \cellcolor{lightblue}29.76 & 34.73 & 4.27 & 26.27 & 27.35 & 42.29 & 21.43 & 29.68 & 35.61 \\
InternVL3.5-8B & \cellcolor{lightblue} 30.96 & 32.34 & 8.55 & 26.27 & 32.49 & 38.83 & 33.33 & 28.39 & 33.45 \\
MiniCPM-V-4.5 & \cellcolor{lightblue}34.36 & 37.72 & 15.38 & 28.11 & 32.84 & 45.48 & 35.71 & 39.35 & 38.13 \\
LLaVA-OneVision-1.5-8B & \cellcolor{lightblue}31.64 & 36.53 & 2.56 & 32.38 & 26.89 & 46.28 & 33.33 & 31.61 & 34.53 \\
\bottomrule
\end{tabular}
}
\end{table*}

\begin{table*}[!ht]
\centering
\caption{
Evaluation results for the ``Physical Commonsense Probing'' experiment. 
``Avg.'' indicates the overall average accuracy.
}
\label{tab:phy_cot}
\vspace{-2mm}
\rowcolors{2}{gray!8}{white}
\resizebox{\textwidth}{!}{
\begin{tabular}{lcccccccccc}
\toprule
\textbf{Model} & \cellcolor{lightblue}\textbf{Avg.} & \textbf{Appearance Order} & \textbf{Counting} & \textbf{Planning} & \textbf{Relation} &   \textbf{Relative Distance} & \textbf{Relative Size} & \textbf{Relative Speed} & \textbf{Spatial State} \\
\midrule
\rowcolor{lightgrey} \multicolumn{11}{c}{$\bullet$ \textit{Vanilla CoT}} \\
Qwen3-VL-8B & \cellcolor{lightblue}51.52$_{\textcolor{red}{\uparrow 4.00}}$ & 63.47 & 50.43 & 51.53 & 44.39 & 54.52 & 90.48 & 61.29 & 51.80 \\
InternVL3.5-8B & \cellcolor{lightblue}50.36$_{\textcolor{mygreen}{\downarrow 0.16}}$ & 59.28 & 51.28 & 49.29 & 43.02 & 56.65 & 61.90 & 65.16 & 51.08 \\
MiniCPM-V-4.5 & \cellcolor{lightblue}44.88$_{\textcolor{red}{\uparrow 1.08}}$ & 55.09 & 40.17 & 48.68 & 35.47 & 53.19 & 73.81 & 59.35 & 39.93 \\
LLaVA-OneVision-1.5-8B & \cellcolor{lightblue}42.88$_{\textcolor{mygreen}{\downarrow 2.64}}$ & 48.50 & 39.32 & 46.44 & 33.30 & 53.99 & 69.05 & 54.19 & 39.57 \\
\rowcolor{lightgrey} \multicolumn{11}{c}{$\bullet$ \textit{Physics-aware CoT}} \\
Qwen3-VL-8B & \cellcolor{lightblue}50.80$_{\textcolor{red}{\uparrow 3.28}}$ & 61.08 & 52.14 & 50.51 & 45.88 & 57.71 & 73.81 & 58.06 & 43.17 \\
InternVL3.5-8B & \cellcolor{lightblue}48.24$_{\textcolor{mygreen}{\downarrow 2.28}}$ & 57.49 & 45.30 & 49.29 & 40.39 & 57.98 & 83.33 & 61.29 & 41.01 \\
MiniCPM-V-4.5 & \cellcolor{lightblue}42.52$_{\textcolor{mygreen}{\downarrow 1.28}}$ & 56.29 & 43.59 & 46.64 & 34.67 & 46.81 & 66.67 & 52.90 & 35.97 \\
LLaVA-OneVision-1.5-8B & \cellcolor{lightblue}44.04$_{\textcolor{mygreen}{\downarrow 1.48}}$ & 47.90 & 44.44 & 44.60 & 36.84 & 50.53 & 80.95 & 56.13 & 42.09 \\
\bottomrule
\end{tabular}
}
\end{table*}

\section{Benchmark Comparison}
\label{app:bench_compare}
As shown in Table~\ref{tab:benchmark_comparison}, we provide a more detailed comparison between SCP-Bench and representative spatial reasoning benchmarks. 
Benchmarks such as OmniSpatial~\cite{jia2025omnispatial}, Spatial457~\cite{wang2025spatial457}, and EmbSpatial-Bench~\cite{du2024embspatial} are built around image-only settings and assess spatial reasoning in static scenes. 
Although they offer meaningful insights into how models interpret spatial structure, they fail to capture the evolving spatial relationships that arise in dynamic settings.
All-Angles-Bench~\cite{yeh2025seeing} and MMSI-Bench~\cite{yang2025mmsi} extend single-view evaluation to multi-view settings, but they remain confined to image-based static scenes.
Although presented in video form, VSI-Bench~\cite{yang2025thinking} effectively evaluates static indoor scenes rather than genuinely dynamic settings.
VLM4D~\cite{zhou2025vlm4d}, STI-Bench~\cite{li2025sti}, and DSI-Bench~\cite{zhang2025dsi} use videos to introduce dynamic spatial reasoning, but their evaluation remains confined to interpreting the information explicitly present in the visible video frames. 
They disregard both the assessment of causal reasoning and the predictive analysis of unobserved portions of a spatial process, whereas spatial causal prediction addresses these aspects and better reflects the demands of everyday situations.
MindCube~\cite{yin2025spatial} examines a model’s ability to hypothesize unseen spatial states using images and text, but it remains restricted to static scenes and lacks visual spatial dynamics.

In contrast, SCP-Bench focuses on spatial causal prediction in dynamic video settings, requiring models to infer spatial states beyond the visible part. 
In addition, SCP-Bench explicitly incorporates diverse perspective settings, covering both egocentric and exocentric viewpoints as well as single-view and multi-view configurations.

\section{Supplementary Results}
\label{app:add_exp}

\subsection{Extended Causal Consistency Experiment}
\label{app:extend_causal_consistency}
To further examine models’ causal consistency in the SCP task, we design an experiment in which the model is required to select the least plausible option instead of the correct one. 
This setting evaluates whether the model can reliably rule out options that contradict the causal structure of the video, providing a complementary perspective on its causal reasoning stability. 

\begin{figure*}[!ht]
  \centering
   \includegraphics[width=0.95\textwidth]{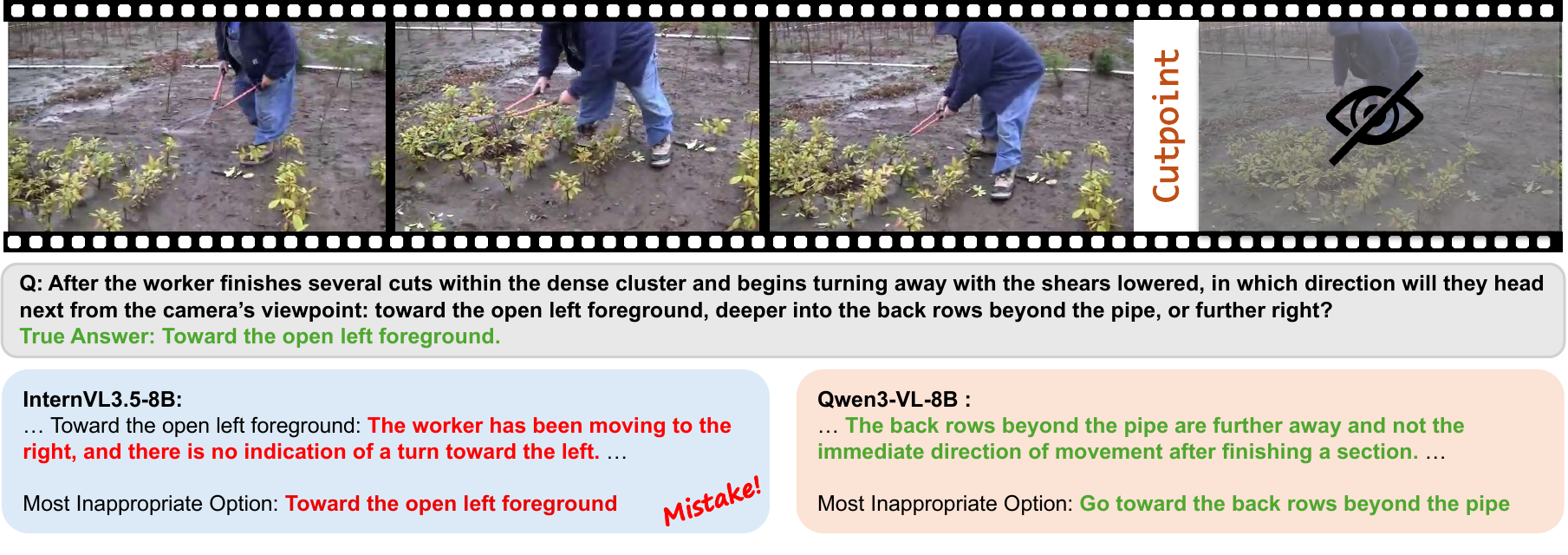}
   \vspace{-2mm}
   \caption{
   Analysis sample for extended causal consistency experiment.
   The models are required to identify the least plausible option, while
   InternVL3.5-8B does not independently assess the plausibility of each choice but instead derives the least plausible option by reasoning backward from its own incorrect answer, revealing a lack of internal causal consistency.
   }
   \label{fig:extended_causal_consistency}
\end{figure*}

As shown in Table~\ref{tab:extended_causal_consistency}, we evaluate Qwen3-VL-8B, InternVL3.5-8B, MiniCPM-V-4.5, and LLaVA-OV-1.5-8B by measuring how often they mistakenly classify the correct answer as the least plausible option, where a lower rate indicates stronger causal consistency.
Although InternVL3.5-8B achieves better performance in the baseline setting, Qwen3-VL-8B produces fewer such errors in this consistency test.
As illustrated in Fig.~\ref{fig:extended_causal_consistency}, Qwen3-VL-8B evaluates each option’s plausibility more independently, whereas InternVL3.5-8B often commits to a presumed correct answer and then selects the option most opposed to it.
When its initial guess is incorrect, this strategy makes it more likely to mislabel the true answer as the least plausible.

\subsection{Physical Commonsense Probing Experiment}
\begin{figure*}[!t]
\centering
\includegraphics[width=0.95\textwidth]{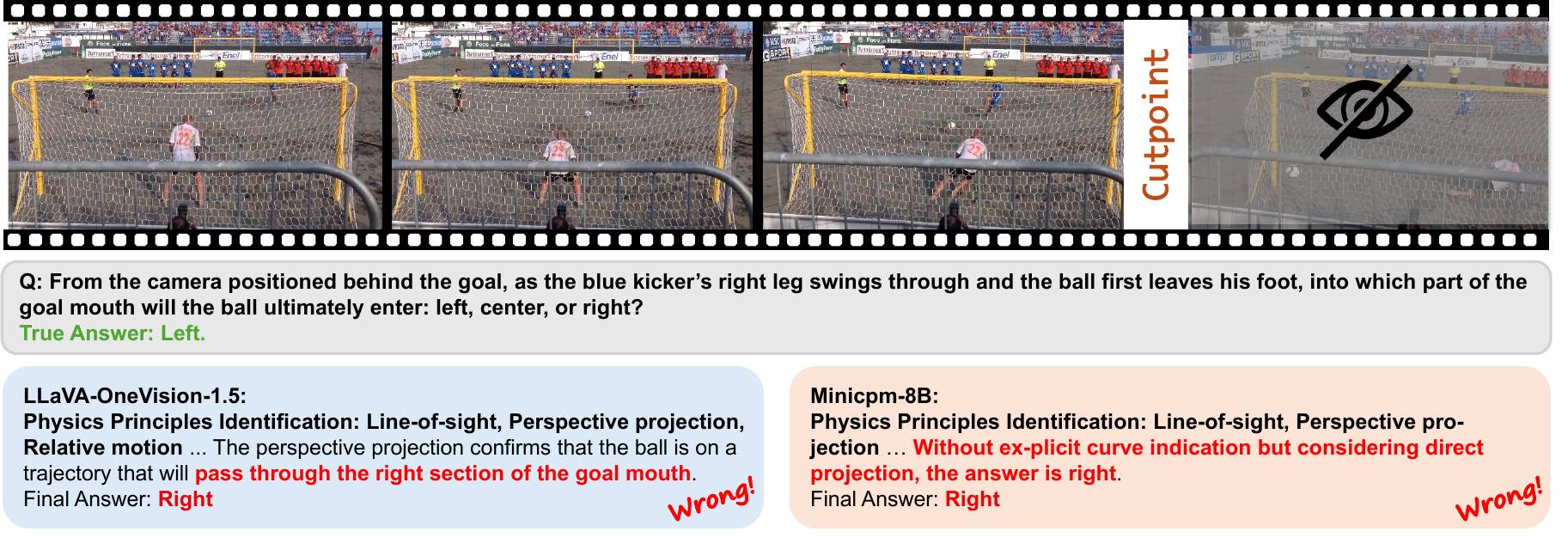}
\vspace{-2mm}
\caption{
Visualization of the physical commonsense probing experiment.
To answer each question, models must first infer the underlying physical principles depicted in the video. Although current models can often recognize the appropriate physical laws, they still struggle to apply them effectively when generating correct predictions.
}
\label{fig:phy_cot}
\end{figure*}

To examine the role of physical commonsense in SCP performance, we introduce a physics-aware CoT prompting strategy. 
The prompt instructs the model to first enumerate the relevant physical principles and then apply them to answer the question. 
However, as shown in Table~\ref{tab:phy_cot}, physics-aware CoT does not clearly improve performance compared to vanilla CoT. 
Analyzing the model’s reasoning traces in Fig.~\ref{fig:phy_cot}  reveals a consistent pattern: although the model can correctly identify the appropriate physical laws, it often fails to apply them to the scene. 
This indicates that current models lack the ability to operationalize physical principles in context and to integrate them into reliable causal reasoning.

\subsection{Error Case Analysis}
\label{app:case}
\begin{figure*}[!t]
  \centering
   \includegraphics[width=0.95\textwidth]{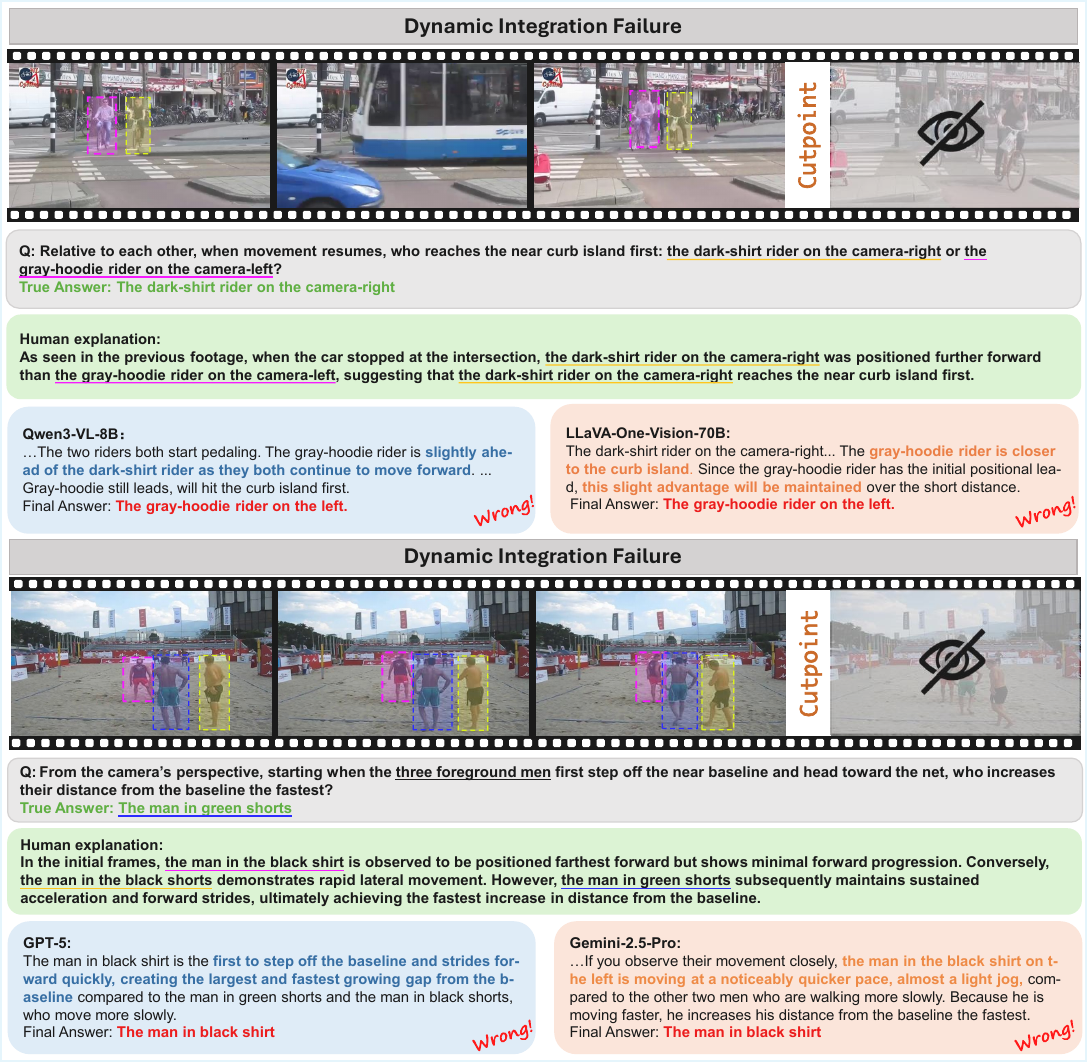}
   \vspace{-2mm}
   \caption{
   Cases of dynamic integration failure. 
   The models fail to perceive dynamic changes and remain confined to static observations.
   }
   \label{fig:error_case1}
\end{figure*}

\begin{figure*}[!t]
  \centering
   \includegraphics[width=0.95\textwidth]{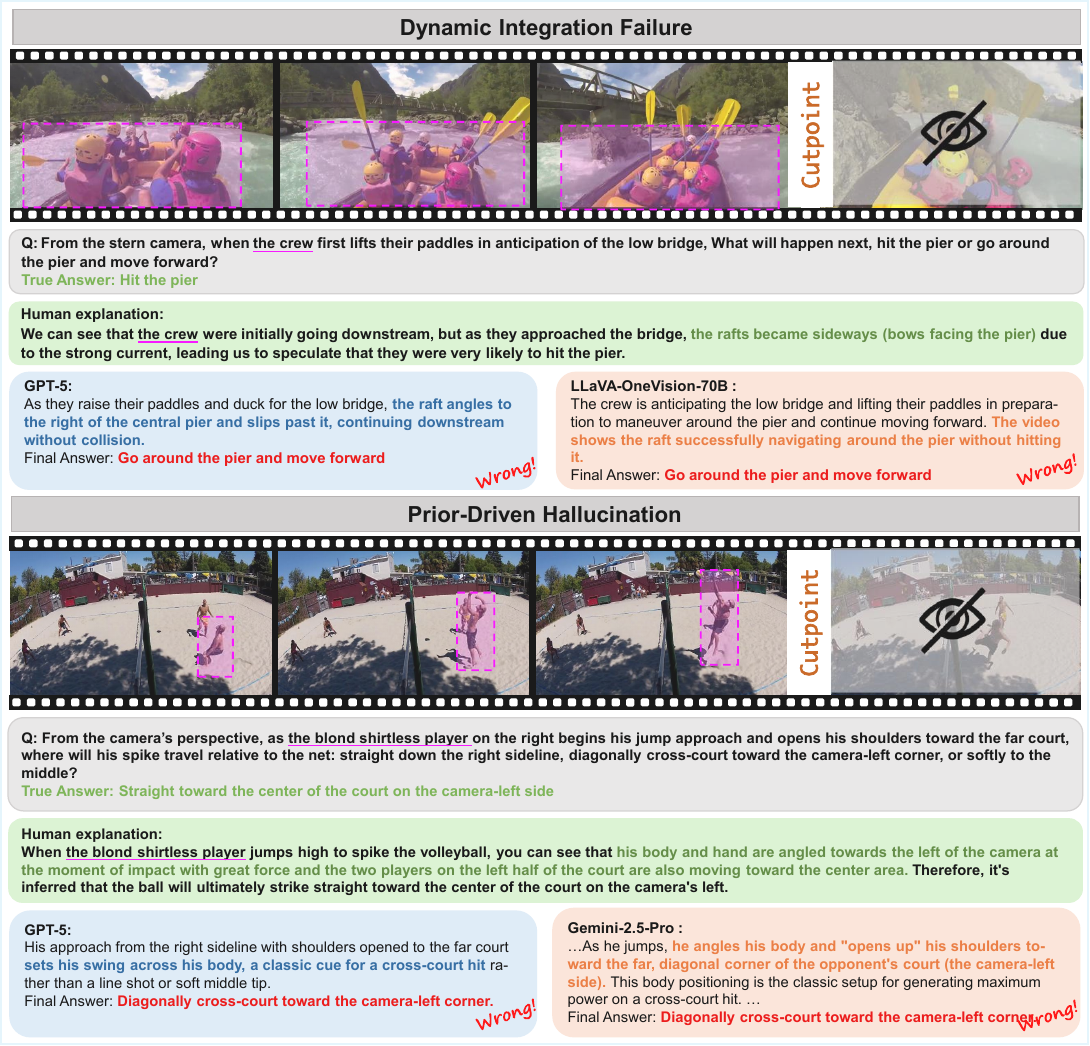}
   \vspace{-2mm}
   \caption{Cases of dynamic integration failure and prior-driven hallucination. The models struggle to perceive and infer the correct global spatial dynamics and often overlook critical scene details during reasoning due to prior-driven biases.}
   \label{fig:error_case2}
\end{figure*}

\begin{figure*}[!t]
  \centering
   \includegraphics[width=0.95\textwidth]{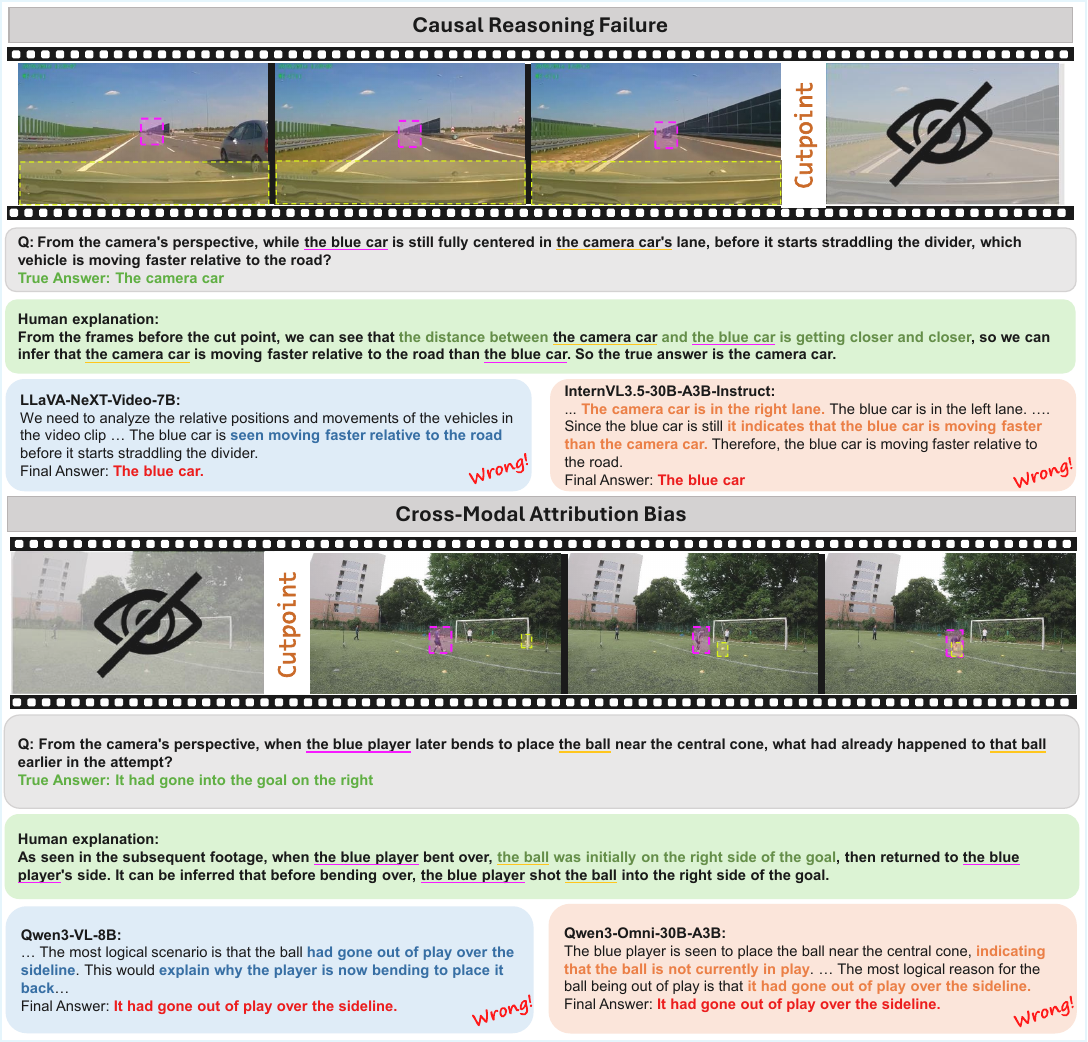}
   \vspace{-2mm}
   \caption{Cases of causal reasoning failure and cross-modal attribution bias. 
   The models struggle to perform causal reasoning and instead produce unsupported conclusions, leading to erroneous predictions. 
   Moreover, they may overlook information from other modalities and fall into single-modality reasoning.}
   \label{fig:error_case3}
\end{figure*}

To further understand why models struggle on the SCP task, we analyze representative error cases and identify several underlying causes of failure:

\begin{compactitem}
\item \textbf{Dynamic Integration Failure.} 
A common failure cause is the model’s inability to integrate motion cues over time, leading it to rely on isolated frames or short local motion fragments instead of coherent temporal dynamics.
As illustrated in Fig.~\ref{fig:error_case1}, the first case considers only the riders’ initial positions while ignoring the subsequent motion changes. 
The second case focuses solely on the individuals’ current distances, overlooking how their relative positions evolve as they move.
In the first case in Fig.~\ref{fig:error_case2}, the model captures only the boat’s local turning motion but fails to recognize its global dynamic trajectory.

\item \textbf{Prior-Driven Hallucination.}
The model fails to attend to detailed visual cues and instead leans too heavily on its prior assumptions.
As shown in the second case of Fig.~\ref{fig:error_case2}, although it correctly identifies the main evolving motion, it subsequently jumps to a conclusion driven by prior assumptions, ignoring the actual scene details and producing an incorrect answer.

\item \textbf{Causal Reasoning Failure.}
The model fails to produce grounded causal reasoning and instead jumps to premature, unsupported conclusions. 
In the first case of Fig.~\ref{fig:error_case3}, for example, it asserts that the blue car is moving faster without any justification, overlooking the causal reasoning needed to compare how the distance between cars evolves over time, which is a process humans naturally perform.

\item \textbf{Cross-Modal Attribution Bias.} 
The model fails to integrate visual and textual modalities into a coherent line of reasoning, often overcommitting to one source and producing biased conclusions. 
As shown in the second case of Fig.~\ref{fig:error_case3}, it becomes fixated on interpreting the textual mention of the player placing the ball and derives its answer from that alone, while entirely ignoring the visual evidence of the ball’s actual trajectory that should have informed the reasoning.

\end{compactitem}

\clearpage
\twocolumn